\documentclass[10pt,journal,compsoc]{IEEEtran}
\usepackage{hyperref}       % hyperlinks
\usepackage{url}            % simple URL typesetting
\usepackage{booktabs}       % professional-quality tables
\usepackage{amsfonts}       % blackboard math symbols
\usepackage{nicefrac}       % compact symbols for 1/2, etc.
\usepackage{microtype}      % microtypography
\usepackage{xcolor}         % colors
\usepackage{float}
\usepackage{amsmath}
\newtheorem{myDef}{Definition}
\usepackage{graphicx}  %插入图片
\usepackage{bm}
\usepackage[linesnumbered,ruled]{algorithm2e}
\usepackage{subfigure}
\usepackage{makecell}
\usepackage{wrapfig}
\usepackage{stfloats}
\usepackage{bbm}
\usepackage{multirow}
\usepackage{makecell}
\usepackage{colortbl}
\usepackage{cite}

\begin{document}

\title{Steering Graph Neural Networks with Pinning Control}

\author{Acong Zhang, Ping Li, Guanrong Chen,~\IEEEmembership{Life Fellow,~IEEE} \protect\\
\IEEEcompsocitemizethanks{\IEEEcompsocthanksitem Acong Zhang and Ping Li are with School
of Computer Science, Southwest Petroleum University, Chengdu,
China, 610500. E-mail: dping.li@gmail.com (Ping Li, the corresponding author.). 
\IEEEcompsocthanksitem Guanrong Chen is with the Department of Electrical Engineering, City
University of Hong Kong, Hong Kong, China.
}% <-this % stops an unwanted space
}

% The paper headers
\markboth{Journal of \LaTeX\ Class Files,~Vol.~14, No.~8, August~2015}%
{Shell \MakeLowercase{\textit{et al.}}: Bare Demo of IEEEtran.cls for Computer Society Journals}

\IEEEtitleabstractindextext{%
\begin{abstract}
In the semi-supervised setting where labeled data are largely limited, it remains to be a big challenge for message passing based graph neural networks (GNNs) to learn feature representations for the nodes with the same class label that is distributed discontinuously over the graph. To resolve the discontinuous information transmission problem, we propose a control principle to supervise representation learning by leveraging the prototypes (i.e., class centers) of labeled data. Treating graph learning as a discrete dynamic process and the prototypes of labeled data as “desired” class representations, we borrow the pinning control idea from automatic control theory to design learning feedback controllers for the feature learning process, attempting to minimize the differences between message passing derived features and the class prototypes in every round so as to generate class-relevant features. Specifically, we equip every node with an optimal controller in each round through learning the matching relationships between nodes and the class prototypes, enabling nodes to rectify the aggregated information from incompatible neighbors in a graph with strong heterophily. Our experiments demonstrate that the proposed PCGCN model achieves better performances than deep GNNs and other competitive heterophily-oriented methods, especially when the graph has very few labels and strong heterophily.
\end{abstract}

}

\maketitle
\IEEEdisplaynontitleabstractindextext
\IEEEpeerreviewmaketitle

\IEEEraisesectionheading{\section{Introduction}\label{sec:introduction}}
\IEEEPARstart{G}{raph} or network is widely used for describing the interactions between elements of a complex system, such as those in social networks~\cite{tu2022viral}, knowledge graphs~\cite{bastos2021recon}, molecular graphs~\cite{jin2020hierarchical}, and recommender systems~\cite{lin2022improving}. To deal with those non-Euclidean data for various graph analytical tasks such as node classification~\cite{wang2020nodeaug} and link prediction~\cite{li2019structure}, graph neural networks (GNNs)~\cite{2017Inductive,xu2018powerful} have been developed and shown having superior performances. 

The core of current GNNs such as GCN~\cite{2016Semi} is message passing. In  message passing, feature representations are learned for each node by recursively performing aggregation and transformation on the representations of its immediate neighbors, revealing that information about long-distance neighbors can be captured this way. However, it is still challenging for the labeled nodes to propagate their information far away using a conventional message passing algorithm, since the influence of labeled nodes decays as the topological distance increases~\cite{xu2018representation}. Moreover, increasing message passing number will lead to oversmoothing~\cite{xu2018representation,li2018deeper}, i.e., the case where representations are determined by the graph structure. While techniques like residual connections used in GCNII~\cite{chen2020simple} allow the network architecture to be deeper, they substantially increase the number of learnable parameters and computational complexity of the GNN. 

Another shortcoming of message passing is its negative smoothing effect in the circumstances where the nodes of the same type discontinuously distributed in the topology space. For instance, in heterophilious graphs, the immediate neighbors of a node come from different classes. It has been revealed~\cite{zhu2020beyond} that in smoothing such nodes, message passing forcefully make the feature representations of the nodes with different labels approximate the average of the local neighborhood, thus deteriorating the representation learning. Previous work~\cite{abu2019mixhop,chen2020simple,zhu2020beyond} suggests some solutions by improving the aggregation scheme using intermediate representations (e.g., residual connections) or higher-order neighborhood, but they can be sub-optimal. These methods either leverage the statistics about homophily of a graph while neglecting the differences of the homophily level between nodes, or depend on stacking more convolution layers, as commonly observed.

In this paper, we address these questions by proposing a novel principle that enhances supervision for message passing with the existing labeled data. Intuitively, the representation of a node should be close to the representation of the prototype of the class it belongs, besides its local neighbors. Thus, the class prototypes (i.e., class centers) of the labeled nodes are ideal references for node representation learning. In other words, the class prototypes can be used to supervise node representation learning. In this work, we propose a strategy to implement the class prototype supervised message passing.

Our idea comes from pinning control of complex networks~\cite{DeLellis2010}, where the coupled nodes are dynamic variables and a certain number of controllers are “pinned” (i.e. exerted) to some nodes to regulate the behaviors of all agents towards a desired common state. Here, a controller is a control feedback scheme, which alters the difference between a dynamical variable and a desired state so that the variable will asymptotically approach the desired state. Inspired by the pinning control idea, we consider the feature representation learning as a discrete dynamic process and the representations of class prototypes in training data as the “desired states”, with which we design the controllers to regulate node representations based on the differences between the current node representations and the “desired class representations”. In this process, the class prototypes play the role of supervision.

Different from pinning control in complex systems where the problem is to decide the minimum number of nodes needed for achieving global synchronization, our goal is to infer class labels for all unlabelled nodes, while allowing all nodes to be “pinned” (i.e., supervised). A challenge in applying pinning control to GNNs is which “controllers” will be used to pin which nodes. This is unknown beforehand, because one controller is associated with only a certain class (corresponding to one desired class prototype representation). Ideally, each node should be supervised by one controller associated with the class of that node, but this is impossible for those nodes whose labels are invisible. To resolve this issue, we propose a dynamic pinning control method, which learns the matching relationships between nodes and class prototypes (i.e., a set of “desired states”) each time when message passing is performed. This way, the pinning control can be adjusted adaptively so as to better align the desired states and the pinned nodes in each graph convolution iteration. By steering the message passing with class prototype-based pinning control, it is able to teleport the information about classes to the regions that are weakly influenced by the labeled nodes without resorting to deep architectures. Meanwhile, the feedback from the controllers allows message passing to rectify the noisy information aggregated from the incompatible\footnote{Following previous work~\cite{zhu2020beyond}, we use compatibility to indicate whether two connected nodes have the same class label, thus two connected nodes with different labels are called incompatible. The overall compatibility of the connected pairs is measured by the homophily index. That is, two connected nodes in a homophilous graph are more likely to be compatible than those in a heterophilious graph.} neighbors of the central node in a heterophilious graph. Thus, the proposed dynamic pinning control differs from the conventional pinning control in "pinning" the class labels rather than pinning the nodes. We experimentally verify these by comparing it with the vanilla message passing-based GCN and the state-of-the-art GNNs for the task of semi-supervised node classification across the full spectrum of heterophily. 

In summary, the main contributions of our work are as follows:
\begin{itemize}
	\item We propose a novel graph representation learning framework by introducing the methodology of pinning control into message passing, which uses learning feedback controllers to supervise representation learning towards the representations of class prototypes so to transmit the class-relevant information to each node directly.
	
	\item We develop an end-to-end model to learn the representations of class prototypes and dynamically select a pinning controller for each node and update the pinning control relationships adaptively during message passing, which enables unlabeled nodes to be supervised by the prototype of the latent class directly, solving the problem of distant message passing.

	\item We conduct extensive experiments on a variety of real-world graph datasets demonstrate that the proposed method improves the performance of the vanilla message passing GCN by a large margin and generally outperforms the state-of-the-art GNN models with different message passing schemes, especially when the network has limited labels.
	
\end{itemize}

\section{Preliminary}

\begin{figure*}[htbp]
	\centering
	{
	\subfigure
    { 
    \centering
    \includegraphics[scale=0.31]{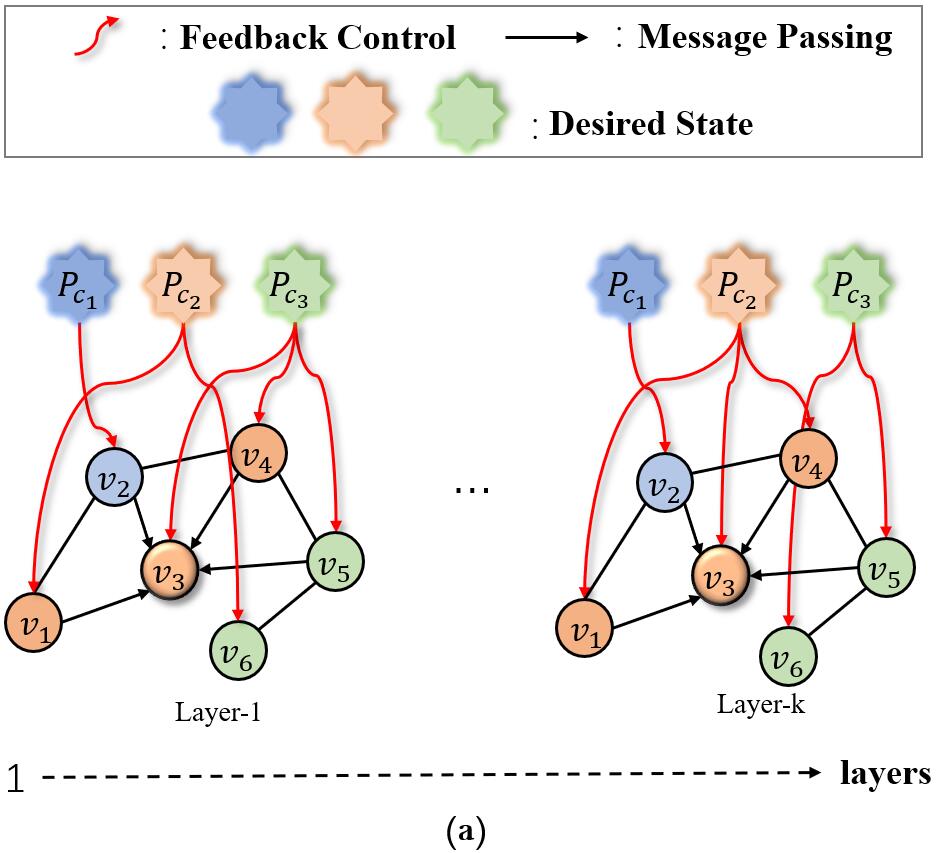}}
	\hspace{0.2cm}
	\subfigure
	{
    \includegraphics[scale=0.30]{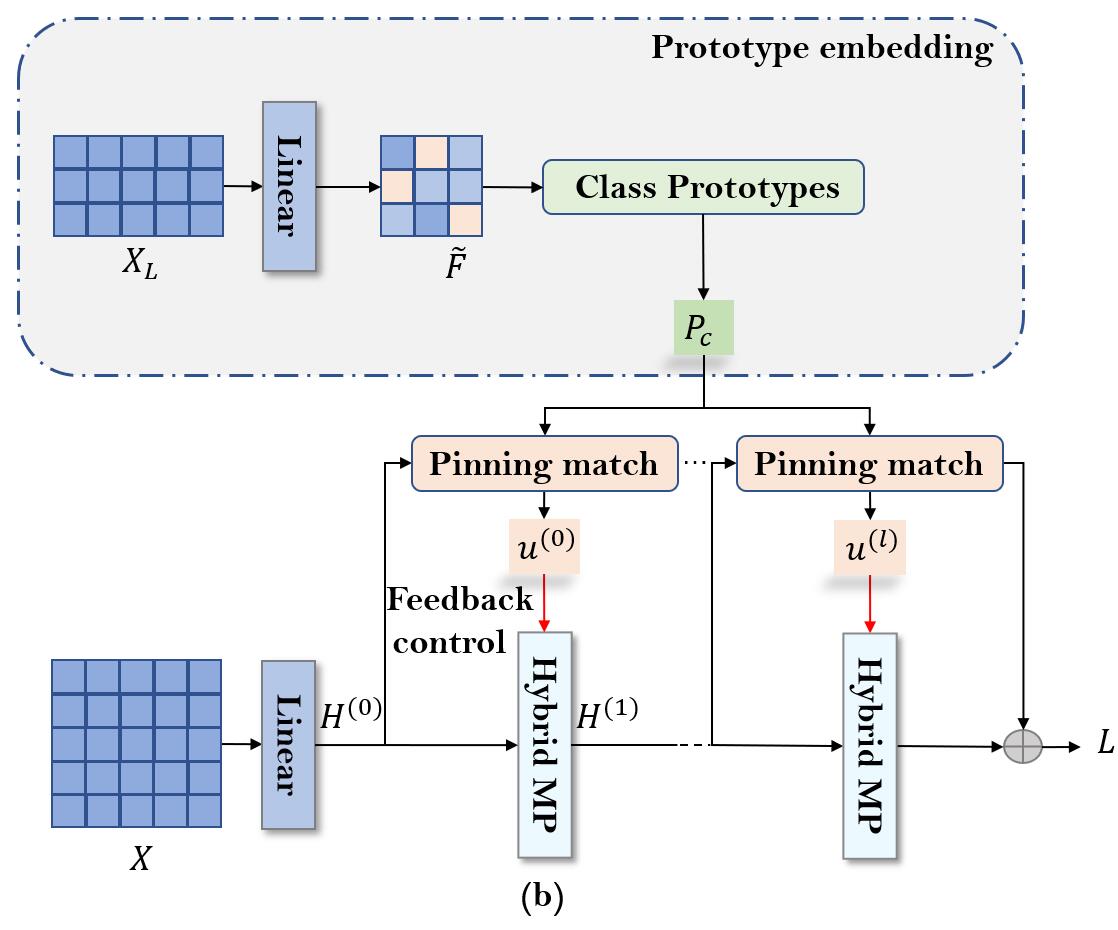}}
	\caption{ (a) Pinning control principle for graph neural networks. (b) The instantiation of pinning controlled message passing based on learnable desired states (i.e., class prototypes), where a feedback controller $u_{i}^{(l)}$ for any node $i$ in the $l$-th layer is implemented by the difference between its current representation and the matched prototype.}
	\label{fig:Model}
}
\end{figure*}

\noindent\textbf{Notation}. Consider an undirected and unweighted graph $\boldsymbol{G = (V,E)}$ with $f$-dimensional attributes $\boldsymbol{X} \in\mathbb{R}^{n\times f}$ on the nodes, where $V$ is the set of nodes, $\boldsymbol{E}$ is the set of edges, and $n=\left |\boldsymbol{V} \right |$ is the number of nodes. The adjacency matrix associated with graph $G$ is denoted as $\boldsymbol{A} \in\mathbb{R}^{n\times n}$. Let $\boldsymbol{D}$ be the diagonal  degree matrix. Then, the graph with self-loop at every node can be represented as $\boldsymbol{\widetilde{A}} = \boldsymbol{A} + \boldsymbol{I_{n}}$, and the corresponding diagonal degree matrix is $\boldsymbol{\widetilde{D}} = \boldsymbol{D} + \boldsymbol{I_{n}}$. Thus, the self-looped adjacency can be symmetrically normalized as $\boldsymbol{\widehat{A}} = \boldsymbol{\widetilde{D}^{-1/2}\widetilde{A}\widetilde{D}^{-1/2}}$. In this work, we focus on semi-supervised node classification~\cite{chapelle2009semi,van2020survey}, which trains a classifier $f_{\theta }(\cdot )$ on the labeled node set $\boldsymbol{T}$ to predict class labels $y$ for the unlabeled node set $\boldsymbol{U}=\boldsymbol{V}-\boldsymbol{T}$. We denote the training sets of different classes by $(C_1,C_2,...,C_c)$, where $c$ is the number of classes. 

\noindent\textbf{Homophily and Heterophily.} As one of the graph properties, homophily means that the connected node pairs tend to have similar features and belong to the same class. Conversely, the connected node pairs are less similar in heterophily. We measure homophily-heterophily using different relationships between node labels and graph structure. There are two commonly ways to measure homophily: edge homophily~\cite{zhu2020beyond} and node homophily~\cite{pei2020geom}, defined as follows:
\begin{myDef}
(\textbf{Homophily Ratio}). Given a graph $G$, the homophily ratio $\widetilde{h} = \frac{\left | \left \{ (v_i, v_j)\mid (v_i, v_j)\in E \wedge y_i = y_j\right \}\right |}{\left | E\right |}$, where $(v_i, v_j)$ is the intra-class edge.
\end{myDef}
\begin{myDef}
(\textbf{Node-level Homophily Ratio}). Investigating homophily on graphs from a local perspective, the node-level homophily ratio is defined as $\widehat{h}_i = \frac{\left | \left \{ v_j\mid v_j\in \mathcal{N}(v_i) \wedge y_i = y_j\right \}\right |}{\left | \mathcal{N}(v_i)\right |}$.
\end{myDef}

The homophily is strong if the homophily ratio value close to 1, while the heterophily is strong if homophily ratio value close to 0.
 
\noindent\textbf{Message Passing GNN.} Learning a representation vector of a node $h_v$ from the graph structure and node attributes $\boldsymbol{X}$ lies in the core of GNNs. Modern GNNs follow message passing in a neighborhood to approximate graph convolutions, where one can iteratively update the representation of a node by aggregating representations of its neighbors. After $k$ times of message passing, the structural information within its $k$-hop neighborhood can be captured by the node representation. Formally, the $k$-th step of message passing in a GNN is
\begin{equation}
\begin{split}
    & a_v^{(k)}  =   AGGREGATE^{(k)}(\{ h_u^{(k-1)}: u\in \mathcal{N}(v)\}), \\
    & h_v^{(k)}  = COMBINE(h_v^{(k-1)}, a_v^{(k)}),
\end{split}
\end{equation}
where $h_v$ is the representation of node $v$ in the $k$-th layer and $\mathcal{N}(v)$ is a set of nodes directly connected to node $v$. In initialization, $h_v^{(0)} = \boldsymbol{X}_v$. By choosing the element-wise \textit{mean} pooling of the neighborhood $\mathcal{N}(v)$ as $AGGREGATE^{k}(\cdot)$ function and summing as $COMBINE^{k}(\cdot)$, the vanilla GCN~\cite{2016Semi} can be formulated as the integration of two functions:

\begin{equation}\label{eq:GCN_update}
	\boldsymbol{H}^{(l+1)} = ReLU(\boldsymbol{\widehat{A}H}^{(l)}\boldsymbol{W}^{(l)}),
\end{equation}
where $\boldsymbol{W}^{(l)}$ is a layer-specific trainable weight matrix, which can be learned by minimizing the cross-entropy between ground truth and predicted labels on the training set $\boldsymbol{T}$, as
\begin{equation}\label{eq:cr_loss}
    \mathcal{L}=-\sum_{v_i \in T} \sum_{k=1}^{C} y_{i k} \ln z_{i k},
\end{equation}
in which $z = softmax(\boldsymbol{H})$ is the output of last layer.

It is interesting to take Eq.(\ref{eq:GCN_update}) as a coupled discrete dynamic system, wherein each node represented by its feature vector evolves at every iteration. Therefore, it is possible to adopt control methods to guide the learning process towards some desired states (e.g., class centers in the training data), obtaining class-relevant representations.

\noindent\textbf{Pinning Control.} Our method is inspired by $\emph{Pinning Control}$ of complex networks~\cite{chen2022pinning}, which aims to synchronize a set of coupled nonlinear systems to a desired state $\textbf{x}_s$. The pinning control method to achieve this is to pin or control some of the nodes with a state-feedback law, which is described by
\begin{equation}
   \dot{x}_i(t) = f(x_i(t), t) -  \epsilon\sum_{j\in\mathcal{N}_i} A_{ij}(x_i(t)-x_j(t)) - \delta_iqu_i(t), 
\end{equation}
where $x_i \in \mathbb{R}^d$ is the state vector of node $i$, $f(\cdot)$ describes the node dynamics, $A_{ij}$ defines the adjacency between node $i$ and $j$, i.e., node $i$ and $j$ is connected if its value is $1$, otherwise $0$. Note that only when $\delta_i = 1$, the controller $u_i = x_i(t)-\textbf{x}_s(t)$ is pinned at node $i$ with control gain $q$.

This scheme can be readily extended to discrete-time networked systems as follows:
\begin{equation}\label{eq:disC}
   x_i(k+1) = f(x_i(k)) -  \epsilon\sum_{j\in\mathcal{N}_i} A_{ij}(x_i(k)-x_j(k)) - \delta_iqu_i(k), 
\end{equation}
where $k$ denotes the $k$-th time step. In particular, when there is no control action exerted on any nodes and the coupled units are linear systems, i.e., $f(x_i(k)) = x_i(k)$, the discrete-time networked system can be simplified as: $\boldsymbol{X}(k+1) = (\boldsymbol{I}-\epsilon\boldsymbol{L})\boldsymbol{X}(k)$, where $\boldsymbol{L} = \boldsymbol{D} - \boldsymbol{A}$ is the Laplacian matrix. By letting $\epsilon = 1$ and replacing the Laplacian with the symmetrically normalized Laplacian $\boldsymbol{\widetilde{L}}= \boldsymbol{I}-\boldsymbol{\widehat{A}}$, the coupled system exactly describes a one-layer graph convolution: $\boldsymbol{X}(k+1) = \boldsymbol{\widehat{A}}\boldsymbol{X}(k)$, the compact form of Eq.(\ref{eq:GCN_update}). Then, the augmented term $u_i(k)$ in Eq.(\ref{eq:disC}) can be converted as a regulator to rectify the learned representations by graph convolution, which is fulfilled when the features of incompatible neighbors are aggregated, e.g., message passing over heterophilious graphs. 

\noindent\textbf{Other Related Work.} As the core component of GNNs, message passing was first proposed in MPNN~\cite{gilmer2017neural} to unify various GNN models that leverage message passing algorithms and aggregation procedures on graphs. Among the variants of message passing GNNs, GCN~\cite{2016Semi} uses a linear aggregation function for the combination of the features from the immediate neighbors. Another GNN model that adopts linear aggregation is GAT~\cite{2017Graph}, which learns the attentive weights for aggregating features at each iteration round. More recently, in order to expand the receptive field for the commonly used two-layer GCN models, personalized page rank is used for deep message passing in   APPNP~\cite{klicpera2018predict}. On the other hand, the residual connection technique is borrowed from deep convolutional networks to GNNs for stacking more layers. Other examples include JKNet~\cite{xu2018representation}, GCNII~\cite{chen2020simple}, EGNN~\cite{zhou2021dirichlet}, AP-GCN\cite{spinelli2020adaptive}, NDLS~\cite{zhang2021node}, DAGNN \cite{liu2020towards}, in which residual connections are employed to preserve the node representations at the previous layer and thus alleviate the over-smoothing problem. However, residual connection based deep architectures generally suffer from high computational complexity. It is also noteworthy that in inductive learning setting, there are some nonlinear message passing GNNs, e.g., GraphSAGE~\cite{2017Inductive}  VR-GCN~\cite{chen2017stochastic}, Fast-GCN~\cite{chen2018fastgcn}, Cluster-GCN~\cite{chiang2019cluster}, GraphSAINT~\cite{zeng2019graphsaint}.
\section{Methodology}

\begin{algorithm}[t]
% \SetAlgoLined
    \caption{The process of PCGCN}
    \label{alg:algor}
    \LinesNumbered
    \KwIn{Graph $\boldsymbol{G = (V,E)}$, the original attribute  $X\in\mathbb{R}^{n\times f}$, the number of class $k$, the linear layer $\widehat{g}_{\theta}(\cdot)$, the labeled node set $T$.}
    \KwOut{Predicted label $\widehat{y}$.}
    Parameter initialization;
    \\
    \For{epoch=1,2,$\cdots$, Epochs}{
    // Learn the embedding of class prototypes.
    \\
    \For{j from 1 to k}{
        \eIf{$T_{c}$ is not none}{
        $\boldsymbol{P}_{\boldsymbol{c}_j}\leftarrow$ Eq.(\ref{con:class});}
        {Randomly initialize $P^1\in\mathbb{R}^d,\boldsymbol{P}_{\boldsymbol{c}_j}=P^1$;}
        }
    // Learn node representations
    \\
    The original node representations $\boldsymbol{H}^{(0)}=\widehat{g}_{\theta}(\boldsymbol{X})$
    \\
    \For{$l$=1,2, layers}{
        Obtain the similarity $\boldsymbol{S}^{(l)} \leftarrow $ Eq.(\ref{con:sim});
        \\
        Calculate the relationship $\boldsymbol{B}^{(l)} \leftarrow$ Eq.(\ref{con:rel1}),(\ref{con:rel2});
        \\
        Obtain the node representations $\boldsymbol{H}^{(l)}\leftarrow$  Eq.(\ref{eq:hybrid});
    }
    Calculate labels $\widehat{y}=Classifier(\boldsymbol{H}^{(layers)})$
    \\
    // Optimize
    \\
    Update model parameters to minimize $\mathcal{\widetilde{L}} \leftarrow $ Eq.(\ref{con:loss}).
        \\
        }
\end{algorithm}

Modern message passing GNNs are built on the label consistency assumption that adjacent nodes most probably belong to the same cluster/class. However, this may be risky in some graphs where dissimilar nodes (e.g. nodes with different labels) are more likely to be interconnected. In such a situation, message passing provably fails to capture the incompatibility between connected nodes~\cite{zhu2020beyond}. Moreover, this way of message passing makes it intractable to pass the information about labeled nodes to the long-distant neighbors that are in the same class but located in different regions of the graph. To address these limitations, it is desirable to introduce auxiliary supervision that can directly act on the nodes so as to rectify the misleading message passing between dissimilar nodes. From a control viewpoint, this is analogous to the pinning control of discrete-time networked systems, because the information utilized to supervise representation learning plays a role similar to the controllers in the evolution of the node states. Motivated by this, we propose a pinning control framework on GNNs and introduce an instantiation of the neural control scheme.

Our framework is graphically demonstrated in Figure~\ref{fig:Model}, which contains two types of message passing in each layer of GNNs: neighbor-aggregation based message passing and pinning control based message passing. The later passes the information about how close the current representation is to the representation of a certain class, which will be described in details in the following subsection. In contrast to pinning control of complex networks where a common desired state for all the nodes is known in advance, there is no common and already desired state for the nodes in GNNs. So a challenge for applying pinning control to the semi-supervised graph learning is how to design meaningful “desired states” for the nodes in different classes, and the following question is then how to assign the unlabeled nodes with their desired states.

Here, we present an instantiation to complete this hybrid message passing scheme, as shown in Figure~\ref{fig:Model}(b). The whole architecture consists of three components, namely, \emph{representation learning of the “desired states”}, \emph{matching between the desired states and the pinned nodes}, and \emph{the hybrid message passing based graph convolution layer}. The first component offers the controllers to be applied to driving node representation learning, while the second component is to learn the pinning relationships between the designed states and graph nodes. Then, the feedback control mechanism (i.e., pinning controller) is integrated into the aggregation function in the third component. The details of the above modules are demonstrated in the following subsections.

\subsection{Representation of “Desired States”}
To embody class-relevant information into node representations, we use the class prototypes of the training data to serve as the desired states, which will supervise node representation learning by the learning feedback control.
\begin{myDef}
(\textbf{Class prototype}). Given a graph $G$ and the associated labeled node set $\boldsymbol{T}$, which is partitioned into $c$ classes, namely, $(C_1,C_2,...,C_c)$, a class prototype $\boldsymbol{P}_{\boldsymbol{c}_j}$ is the centroid of the embeddings of all the labeled nodes in class $C_j$.
\end{myDef}

To learn class-relevant representations, the prototype of a class is considered to be the desired representation from representation learning for nodes in that class. This way, the control signal related to a certain class can modify the feature representations of the nodes in this class. 

Intuitively, the original attribute mean of the labeled nodes in the same class can be exploited as the representation of the prototype. The embeddings of prototypes for each class are then defined, as
\begin{equation}
	\boldsymbol{P}_{\boldsymbol{c}_j} = \frac{1}{\left | T_j\right |}\sum_{i\in T_j}^{}g_{\theta}(\boldsymbol{X}_{T_j}(i))\label{con:class},
\end{equation}
where $g_{\theta}(\cdot)$ is a linear layer and $T_j$ represents the set of labeled nodes of class j.
The prototypes convey the information about the corresponding classes, therefore they can serve as the desired states for representation learning. Based on them, it is possible to construct “pinning controllers” to steer the representation learning towards the “desired states”. 
\begin{myDef}
(\textbf{Pinning controller}). Given a graph $G$, a pinning controller for node $i$ refers to a control loop feedback $u_i$, which is the difference of the desired representation and the current representation of node $i$ in the learning process.
\end{myDef}

That is, a pinning controller can be formulated as $u_i = h_i - \boldsymbol{P}_{\boldsymbol{c}_i}$, where ${\boldsymbol{c}_i}$ represents the ground truth class that node $i$ belongs to. However, since most of the node labels are invisible, the “desired state” for node $i$ has to be determined and may not be precisely equipped with the ground truth prototype.  

\subsection{Pinning Control Node Matching}

To address the above question, i.e., which class prototype a node should be associated with in order to obtain the optimal pinning controller, the scheme learns the bipartite matching matrix that depicts the relationships between prototypes and graph nodes (i.e., the connections indicated by the red lines in Figure~\ref{fig:Model}(a)). Specifically, we define the pinning relationship based on the similarity between node feature representation and prototype embedding, as
$\boldsymbol{S}^{(l)}  = \boldsymbol{H}^{(l-1)} \boldsymbol{P_c}^{T}$, where $\boldsymbol{P_c}$ is composed of $c$ prototype embeddings and $l$ refers to the $l$-th round of message passing, suggesting that our method allows the control relation to be adapted to the update of node representations, as depicted in Figure~\ref{fig:Model}(a). Here $\boldsymbol{S}^{(l)}$ reflects how many nodes will be pinned by different class prototypes.

Since we aim to build matching between nodes and the prototypes with the same label, the perfect relation would be that nodes of the same type are pinned by the same prototype, which would favor the separation between different classes. Intuitively, the key to achieve this goal is to align the pinning relationship between node and prototype within the neighborhood of the concerned node, so that neighboring nodes are pinned by similar prototypes. Consequently, the pinning relationship $\boldsymbol{S}^{(l)}$ can propagate on the graph, i.e., $\widehat{A}\boldsymbol{S}^{(l)}$. Note that, an implicit assumption here is that nodes tend to be connected with similar nodes in graphs. However, in some graphs~\cite{zhu2020beyond} (i.e., graphs with weak homophily) nodes are more likely to be adjacent to nodes with different labels. Therefore, it would be better to distance a node from its incompatible neighbors in terms of pinning similarity, i.e., $(I-\widehat{A})\boldsymbol{S}^{(l)}$. For a general graph, we combine the above two quantities to improve the approximation of pinning relationships:
%In order to better adapt to the dataset, we hope that in the homophily dataset, the target node is similar to the matching class prototype of its neighbor node, and in the heterophily data, it should not be similar. Therefore, we use high and low frequency filters to act on similarity, which is described as follow:
\begin{equation}
	\boldsymbol{\widetilde{S}}^{(l)} = \alpha^{(l)}\widehat{A}\boldsymbol{S}^{(l)}+(1-\alpha^{(l)})(I-\widehat{A})\boldsymbol{S}^{(l)}\label{con:sim}
\end{equation}
where $\alpha^{(l)}$ is a learnable parameter in each layer. To obtain the index of the prototype most suitable for a certain node $i$, we normalize the similarities between node $i$ and all $c$ prototypes, and retrieve the index of the prototype that has relatively maximal normalized similarity, which is calculated by
\begin{equation}
\begin{aligned}
	IX^{(l)}(i) &= \arg softmax(\boldsymbol{\widetilde{S}}_{i}^{(l)}) \\
 & =  \arg max_{1\leq j \leq c} \frac{e^{\frac{1}{\tau }\boldsymbol{\widetilde{S}}_{ij}^{(l)} }}{\sum_{r}e^{\frac{1}{\tau }\boldsymbol{\widetilde{S}}_{ir}^{(l)}}}\label{con:rel1},
 \end{aligned}
\end{equation}
where $IX^{(l)}$ is an $n$-dimensional vector that records the prototype index for each node, and $\tau $ is a predefined temperature. A smaller $\tau $ makes a skewed output distribution, so that the influence of large similarity values get amplified. As a result, the vector $IX^{(l)}$ depicts the matchable prototype for each node. We translate this bipartite relationship into a sparse matrix:
\begin{equation}
  \boldsymbol{B}_{ij}^{(l)} = \left\{\begin{matrix}
 1,& IX^{(l)}(i)=j\\ 
 0,& otherwise,
\end{matrix}\right.\label{con:rel2}
\end{equation}
where $\boldsymbol{B}^{(l)} \in \mathbb{R}^{n\times c}$ indicates the bipartite mapping between nodes and prototypes.

\subsection{Hybrid Message Passing}
After matching the prototypes and the pinned nodes, we use the state feedback controller  $\boldsymbol{u}_{i}^{(l)} = \beta(\boldsymbol{H}_{i}^{(l)}-\boldsymbol{B}_{i}^{(l)}\boldsymbol{P_c})$ to regulate the representation of node $i$ ($1\leq i \leq n$) in the $l$-th layer, where $\beta$ is a hyperparameter to tune the impact of the difference between two representations through the learning process, and  $\boldsymbol{B}^{(l)}$ is used to look for the prototype representation optimal to each node from $\boldsymbol{P_c}$. Note that from information transmission viewpoint, the pinning control here can also be considered as another kind of message passing, which propagates the information about a certain class. By combing the vanilla message passing and pinning control based “message passing”, we have the following aggregation function: 

\begin{equation}
	\boldsymbol{H}^{(l+1)} = \sigma(\underbrace{\boldsymbol{\widehat{A}}\boldsymbol{H}^{(l)}\boldsymbol{W}^{(l)}}_{Message\,passing}+\beta(\underbrace{\boldsymbol{H}^{(l)}-\boldsymbol{B}^{(l)}\boldsymbol{P_c})\boldsymbol{W}^{(l)}}_{Pinning\, control})\label{eq:hybrid}.
\end{equation}
% where $\boldsymbol{W}_m^{(l)}$ and $\boldsymbol{W}_p^{(l)}$ are learnable parameters. For the sake of simplicity, we use the same feature transformation parameter for both two ways of message passing, i.e., $\boldsymbol{W}_m^{(l)} = \boldsymbol{W}_p^{(l)}$.

\textbf{Semi-supervised classification.} 
It is noteworthy that in our implementation of pinning control, we put the control on all nodes including labeled nodes. To ensure that the labeled nodes are correctly pinned by the label-associated controller, besides the cross-entropy loss as shown in Eq.(\ref{eq:cr_loss}) on node classification, we add a regularization term to penalize the disagreement between the model prediction and the estimation of class consistency (i.e., the normalized similarity between node representation and class prototype representation). Accordingly, the total loss is
\begin{equation}
	\mathcal{\widetilde{L}} = \mathcal{L} - \lambda \sum_{l} \sum_{v_i \in T} \sum_{k=1}^{C} y_{i k} \ln \widetilde{z}^{(l)}_{i k}\label{con:loss},
\end{equation}
where $\widetilde{z}^{(l)}=softmax(\boldsymbol{S}^{(l)})$ indicates the degree of matching between a node and a controlling prototype. 

\noindent\textbf{Complexity} We compare the time complexity of PCGCN with the standard message passing GNN (i.e., the vanilla GCN). Note that one-layer PCGCN is the combination of message passing and pinning control. First, the time complexity of one-round message passing with feature transformation (i.e., GCN layer) is $\mathcal{O}(|E|d+nd^2)$. Second, the time complexity of generating class prototypes is $\mathcal{O}(|T|fd)$, and the implementation of pinning control is $\mathcal{O}(nd+nd^2+c|E|)$. So, the time complexity of one-layer PCGCN is $\mathcal{O}(|E|(c+d)+|T|fd+nd+nd^2)$, which still linearly depends on the network size $n$. However, it should be noted that the pinning control relationships built between nodes and “controllers” introduces additional space overhead to implement the scheme.

\begin{table}[htbp]
	\centering
	\renewcommand\arraystretch{1.0}
	\setlength\tabcolsep{4.0pt}
	\caption{Statistics of the datasets.}
% 	\caption{The statistics of the datasets. The edge density is defined as $\frac{2m}{n^2}$. The homophily ratio is calculated according to the definition in~\cite{zhu2020beyond}.}
	\label{tb:dataintroduce}
	\begin{tabular}{lcccc}
		\hline
		\textbf{Datasets} & \textbf{Nodes} & \textbf{Edges} & \textbf{Features} & \textbf{Classes} \\
		\hline
		Cora & 2,708 & 5,429 & 1,433 & 7 \\
		CiteSeer & 3,327 & 4,552 & 3,703 & 6\\
		Pubmed & 19,717 & 44,338 & 500 & 3 \\
		Cornell & 183 & 295 & 1703 & 5\\
        Wisconsin & 251 & 499 & 1703 & 5\\
        Texas & 183 & 309 & 1703 & 5\\
		Chameleon & 2277 & 36,101 & 2325 & 5\\
		Squirrel & 5201 & 217,073 & 2089 & 5 \\
		Actor & 7600 & 33,544 & 931 & 5 \\
		Flickr & 89,250 &  449878 & 500 & 7\\
		\hline
	\end{tabular}
\end{table}

\begin{table}[htbp]
	\centering
	\renewcommand\arraystretch{1.0}
	\setlength\tabcolsep{4.0pt}
	\caption{Hyperparameters of PCGCN.}
	\label{hyperparameters1}
	\begin{tabular}{lccccccc}
		\hline
		\textbf{Datasets} & \textbf{dropout} & \textbf{hid} & \textbf{layers} & \textbf{lr} & \textbf{wd} &\textbf{$\lambda$}  & \textbf{$\beta$} \\
		\hline
		Cora & 0.8 & 512 & 2 & 0.001 & 5e-4 & 0.1 & 0.6\\
		CiteSeer & 0.7 & 256 & 2 & 0.01 & 5e-4& 0.1 & 0.6\\
		Pubmed & 0.3 & 256 & 2 & 0.001 & 0.0001 & 1 & 0.5\\
		Cornell & 0.4 & 32 & 1 & 0.05 & 5e-4 & 1 & 5\\
		Wisconsin & 0.2 & 128 & 1  & 0.05 & 5e-4 & 1 & 5\\
		Texas & 0.7 & 256 & 1  & 0.05 & 0.001 & 10 & -3\\
		Chameleon & 0.5 & 64 & 2  & 0.01 & 5e-5 & 10 & -0.2\\
		Squirrel & 0.5 & 64 & 2  & 0.01 & 5e-5 & 1 & -0.1\\
		Actor & 0.1 & 64 & 2 & 0.01 & 5e-5 & 10 & -5\\
		Flickr & 0.6 & 128 & 2 & 0.01 & 5e-5 & 0.1 & -0.1 \\
		\hline
	\end{tabular} 
\end{table}

\begin{table*}[htbp]
	\centering
	\renewcommand\arraystretch{1.3}
	\setlength\tabcolsep{3.5pt}
	\caption{Mean accuracy±stdev over different data splits on the ten datasets. The best performance for each dataset is highlighted in bold and the second best performance is underlined for comparison. “N/A” denotes non-reported results. The dash symbols indicate that it is not able to run the experiments due to memory issue}
	\label{heterophily dataset}
	\scalebox{0.95}{
	\begin{tabular}{lccccccccccc}
		\hline
		\textbf{Dataset} & \textbf{Chameleon} & \textbf{Squirrel} & \textbf{Actor} & \textbf{Texas} & \textbf{Wisconsin} & \textbf{Cornell} & \textbf{Flickr}& \textbf{Cora} & \textbf{Citeseer} & \textbf{Pubmed}& \textbf{Avg.}\\
		\textbf{Hom.ratio $\widetilde{h}$} & 0.23 & 0.22 & 0.22 &  0.2 & 0.1& 0.06 & 0.32 & 0.81 &0.74& 0.8 & \textbf{Rank}\\
		\hline
		MLP &46.93±1.7 & 29.95±1.6 & 34.78±1.2 & 79.19±6.3&83.15±5.7& 79.79±4.2 & 44.32±0.2 & 75.13±2.7 &73.26±1.7 &85.69±0.3& 11.1\\
		GCN &65.92±2.5 & 49.78±2.0 & 27.51±1.2&55.14±5.16 &	51.76±3.06&	60.54±5.3 &49.68±0.45 &86.98±1.27 &	76.50±1.36&	88.42±0.5& 10.9\\
		GAT &65.32±1.9 & 46.79±2.0 & 29.03±0.9&52.16±6.63&	49.41±4.09&	61.89±5.05 &49.67±0.81& 87.30±1.10	&76.55±1.23&86.33±0.48& 11.2\\
		GraphSAGE &58.71±2.3 &41.05±1.1 &34.37±1.3 & 82.70±5.9&81.76±5.6 &75.59±5.2 &50.1±1.3 &86.60±1.8 &75.61±1.6 & 88.01±0.8 & 9.8\\
		MixHop & 60.50±2.53 & 43.80±1.48 & 32.22±2.34 & 77.84±7.73&75.88±4.90 &73.51±6.34 &51.92±0.41 &87.61±0.85 & 76.26±1.33 & 85.31±0.61 & 9.9\\
		GCNII &63.86±3.04 & 36.37±1.6 & 34.40±0.7& 77.57±3.8&	80.39±3.4&	77.86±3.7 & 50.34±0.22 &\textbf{88.37±1.25}&	77.33±1.48&	\underline{90.15±0.43} & 7.1\\
		\hline
		H2GCN-1	&58.84±2.1 & 36.42±1.8 & 35.94±1.3&84.86±6.7	&86.67±4.6&	82.16±4.8&51.76±0.1&86.35±1.6&	76.85±1.5	&88.50±0.6 & 7\\
		H2GCN-2	&59.56±1.8 & 37.90±2.0 & 35.55±1.6 & 82.16±5.2&	85.88±4.3&82.16±6.0 & 52.01±0.1 & 88.13±1.4 &	76.73±1.4&	88.46±0.7& 6.6\\
		Geom-GCN &60.90±2.8 & 38.14±0.92  & 31.63±1.15& 60.18&	67.57&64.12 & N/A & 85.35±1.57&\textbf{78.02±1.15}&	89.95±0.47 & 8.6\\
		FAGCN &45.13±2.2 & 31.77±2.1 & 34.51±0.7 &72.43±5.6& 67.84±4.8 &77.06±6.3 & 49.66±0.6 &  87.87±0.8& 76.76±1.6& 88.80±0.6 & 10.2\\
		GPRGNN &46.58±1.71 & 31.61±1.24  & 34.63±1.22&78.38±4.36& 82.94±4.21 &80.27±8.11 & 51.14±0.1 &  87.95±1.18&77.13±1.67& 87.54±0.38 & 8.8\\
		GGCN &\underline{71.14±1.84}&	55.17±1.58&	\textbf{37.54±1.56} & \underline{84.86±4.55}&	86.86±3.29 &	\underline{85.68±6.63} & - &87.95±1.05	&77.14±1.45&	89.15±0.37 & 2.8\\
		LINKX & 68.42±1.38 & \underline{61.81±1.80}  & 36.10±1.55 & 74.60±8.37 & 75.49±5.72 & 77.84±5.81 & 52.24±0.19 & 84.64±1.13 & 73.19±0.99 & 87.86±0.77 & 8.4\\
		GloGNN & 69.78±2.42 & 57.54±1.39  & \underline{37.35±1.30} &  84.32±4.15 & \underline{87.06±3.53} & 83.51±4.26 & \underline{53.97±0.22} & \underline{88.31±1.13} & \underline{77.41±1.65} & 89.62±0.35 & \underline{2.7}\\
		\hline
		PCGCN & \textbf{74.29±1.9}&	\textbf{65.47±2.4}&	36.43±0.9 &\textbf{85.95±3.9}&	\textbf{87.64±3.7}&	\textbf{85.94±6.1}&\textbf{54.64±0.3} &
		87.65±1.5& 77.40±1.3	&\textbf{90.34±0.4}&\textbf{2.0}\\
		\hline
	\end{tabular} 
	}
\end{table*}

\section{Experiments and Evaluation}

To demonstrate the effectiveness of the pinning control scheme, we evaluate the performance of PCGCN against the state-of-the-art GNNs on several graph benchmark datasets for the semi-supervised node classification task. In particular, we address the following questions:\\
\textbf{Q1} Can pinning control help the vanilla message passing to improve the expressive power of GNNs and thus achieve better performance on heterophilous graphs? \\
\textbf{Q2} Is pinning control effective to propagate information for distant nodes? \\
\textbf{Q3} How does a pinning-controlled GNNs depend on the labels?\\%including number of labels and the availability of labels

We implement the PCGCN scheme based on PyTorch and our code is available online\footnote{The hyperlink will be given after acceptance.}. In what follows, we present the experimental settings, followed by our answers to the above research questions one by one.

\subsection{Experimental Setup}
To answer \textbf{Q1}, we evaluate the node classification performance of the proposed PCGCN and compare it with state-of-the-art heterophily-oriented GNN models on heterophilious graphs. Moreover, we test our model on some benchmark graph datasets with strong homophily that cover a full spectrum of heterophily.\\ 
\textbf{Datasets.} We evaluate the performance of PCGCN model and existing GNNs in node classification on various real-world datasets ~\cite{pei2020geom, rozemberczki2021multi, tang2009social,yang2016revisiting}. We provide their statistics in Table~\ref{tb:dataintroduce}, where we compute the homophily level $\widetilde{h}$ of a graph as the average of $h_i$ of all
nodes $v_i \in V$. For all benchmarks, we use the feature vectors, class labels, and 10 random splits
(48\%/32\%/20\% of nodes per class for train/validation/test\footnote{
(Pei et al., 2019) claims that the ratios are 60\%/20\%/20\%, which are different from the real data splits shared on GitHub.} ) from ~\cite{pei2020geom}.

For the Flickr dataset, where nodes represent images and each edge indicates that two images have some common attributes, whereas features are the descriptions of the images. For a fair comparison with the existing results, we adopt the split in ~\cite{zeng2019graphsaint} (i.e.,50\%/25\%/25\% of nodes per class for train/validation/test) for Flickr partition. We report the mean test accuracy and standard deviation of the 10 replicate results.

% The statistics of all nine datasets are summarized in Table~\ref{tb:dataintroduce}.

\textbf{Baselines.} For the heterophilious datasets, we specifically compare our model with heterophily-oriented methods, namely, two variants of H2GCN (i.e., H2GCN-1 and H2GCN-2)~\cite{zhu2020beyond}, Geom-GCN~\cite{pei2020geom}, FAGNN~\cite{bo2021beyond} and one variant of GCNII~\cite{chen2020simple} wherein parameters are shared between layers, GPRGNN~\cite{2021Adaptive}, GCGCN~\cite{2021Two}, LINKX~\cite{LINKX}, GloGNN~\cite{GloGNN}. 
We also compare our scheme with following methods, some of which are shown to be competitive on various graphs: Multilayer Perceptron (MLP), SGC~\cite{wu2019simplifying}, Graph Convolutional Network (GCN)~\cite{2016Semi}, Graph Attention Network(GAT)~\cite{2017Graph},  Mixhop~\cite{abu2019mixhop} and GCNII~\cite{chen2020simple}. 

\textbf{Model Setting.} We implement the proposed PCGCN and some necessary baselines using PyTorch and PyTorch Geometric, a library for deep learning on irregularly structured data built upon PyTorch. We try our best to provide a rigorous and fair comparison between different models. To mitigate the effects of randomness, we run each method $10$ times and report the average performance. For the baseline methods, whose results on the benchmark datasets are publicly available, we directly present the results. For the models without publicly reported results, we use the original codes published by their authors and fine-tune them. All experiments are implemented in PyTorch on 2 NVIDIA RTX3090 24G GPUs with CUDA 11.1. We use Python 3.9.7 and python packages PyTorch 1.8.1, PYG 1.6.3 (cuda 11.1). Table~\ref{hyperparameters1} summarizes the training configuration of PCGCN for semi-supervised node classification, where lr is the learning rate, hid is the hidden dimension, wd is the weight decary, $\lambda$ is the regularization factor and $\beta$ is the control gain.

\subsection{Experimental Results}

\begin{figure*}[htbp]
    \centering
    {
    \subfigure[Chameleon]
    {
    \includegraphics[scale=0.21]{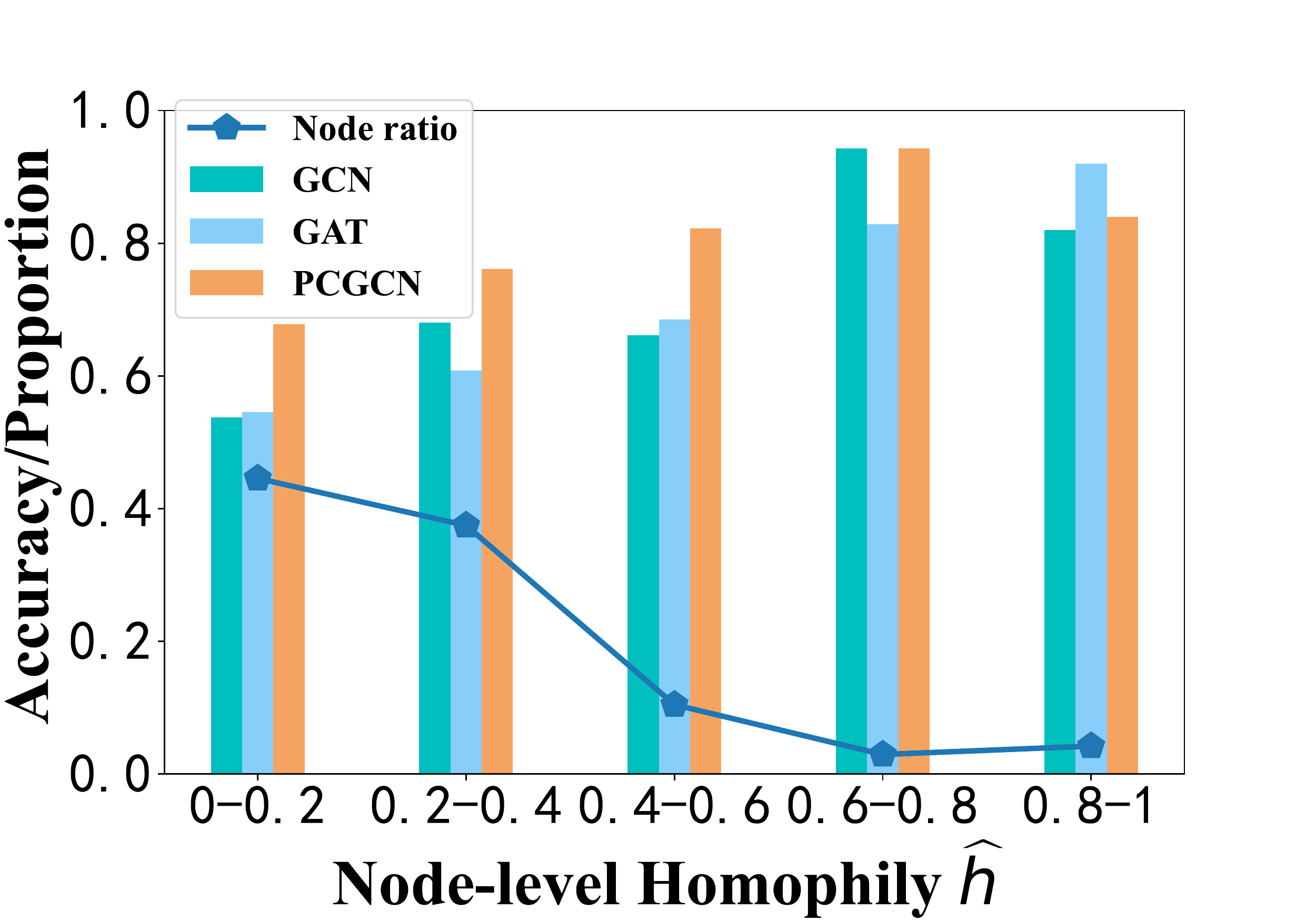}
    }
    \hspace{-0.5cm}
    \subfigure[Squirrel]
    {
    \includegraphics[scale=0.21]{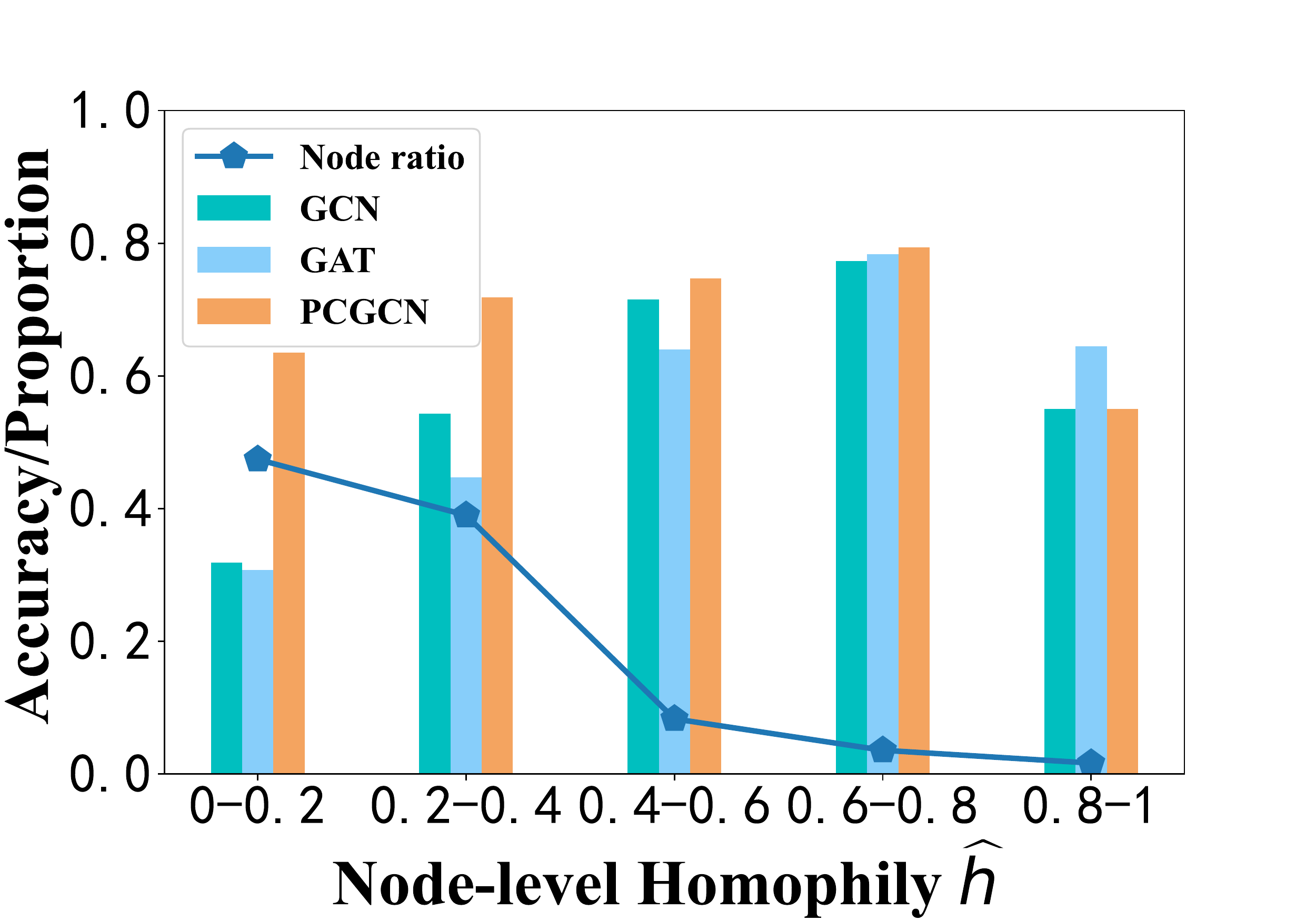}
    }
    \hspace{-0.5cm}
    \subfigure[Actor]
    {
    \includegraphics[scale=0.21]{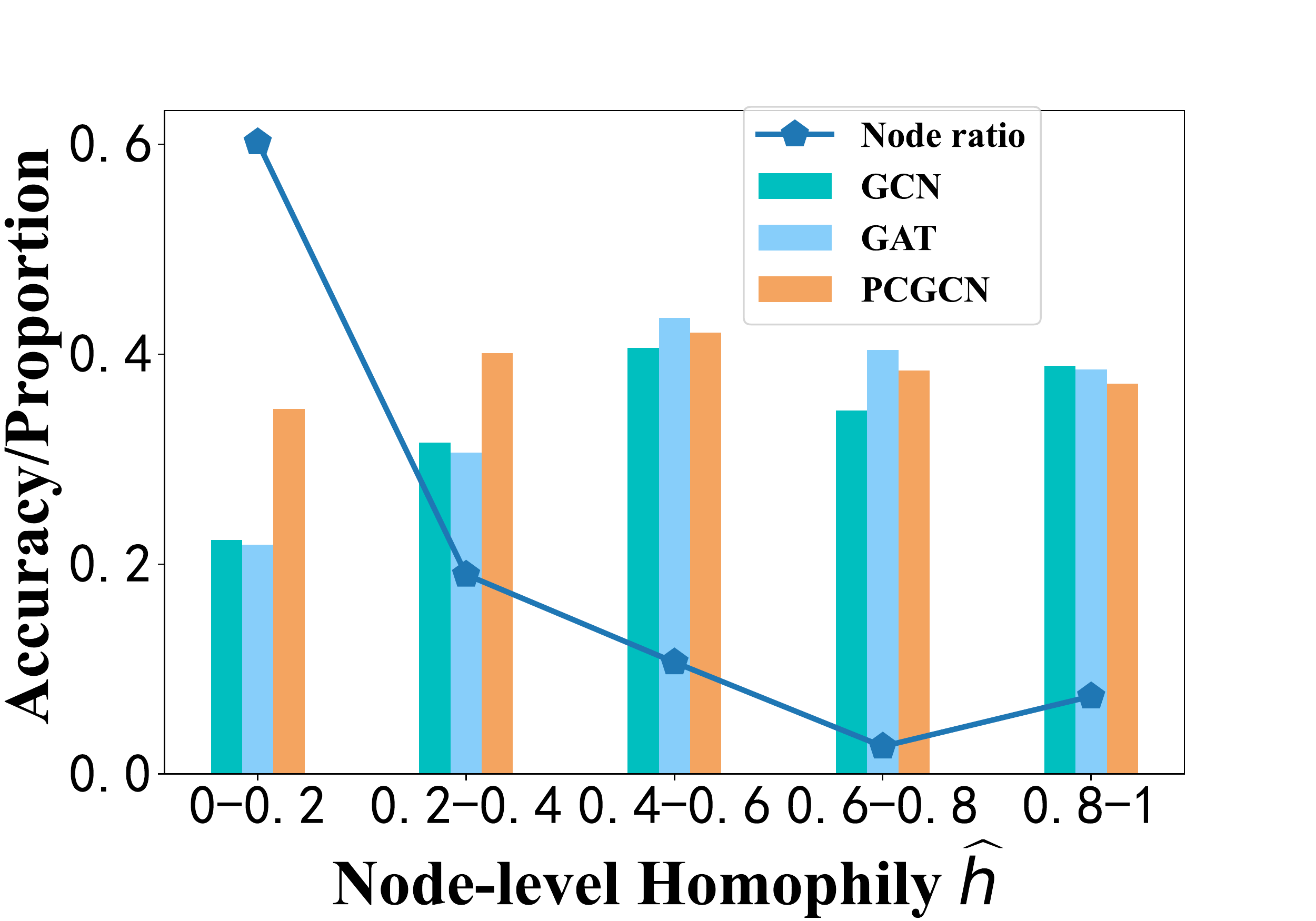}
    }
    \hspace{-0.5cm}
    \subfigure[Texas]
    {
    \includegraphics[scale=0.21]{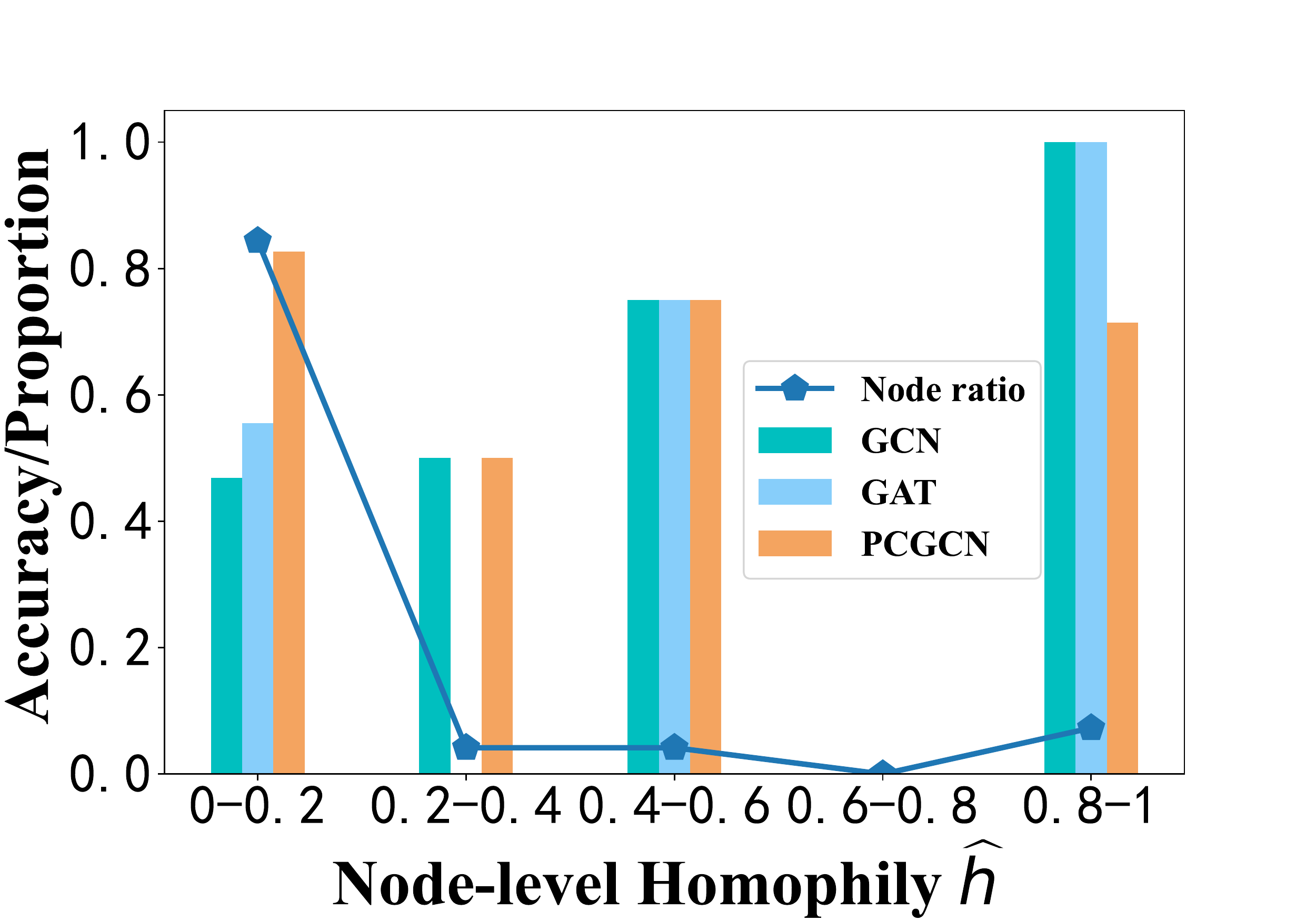}
    }
    \hspace{-0.5cm}
    \subfigure[Wisconsin]
    {
    \includegraphics[scale=0.21]{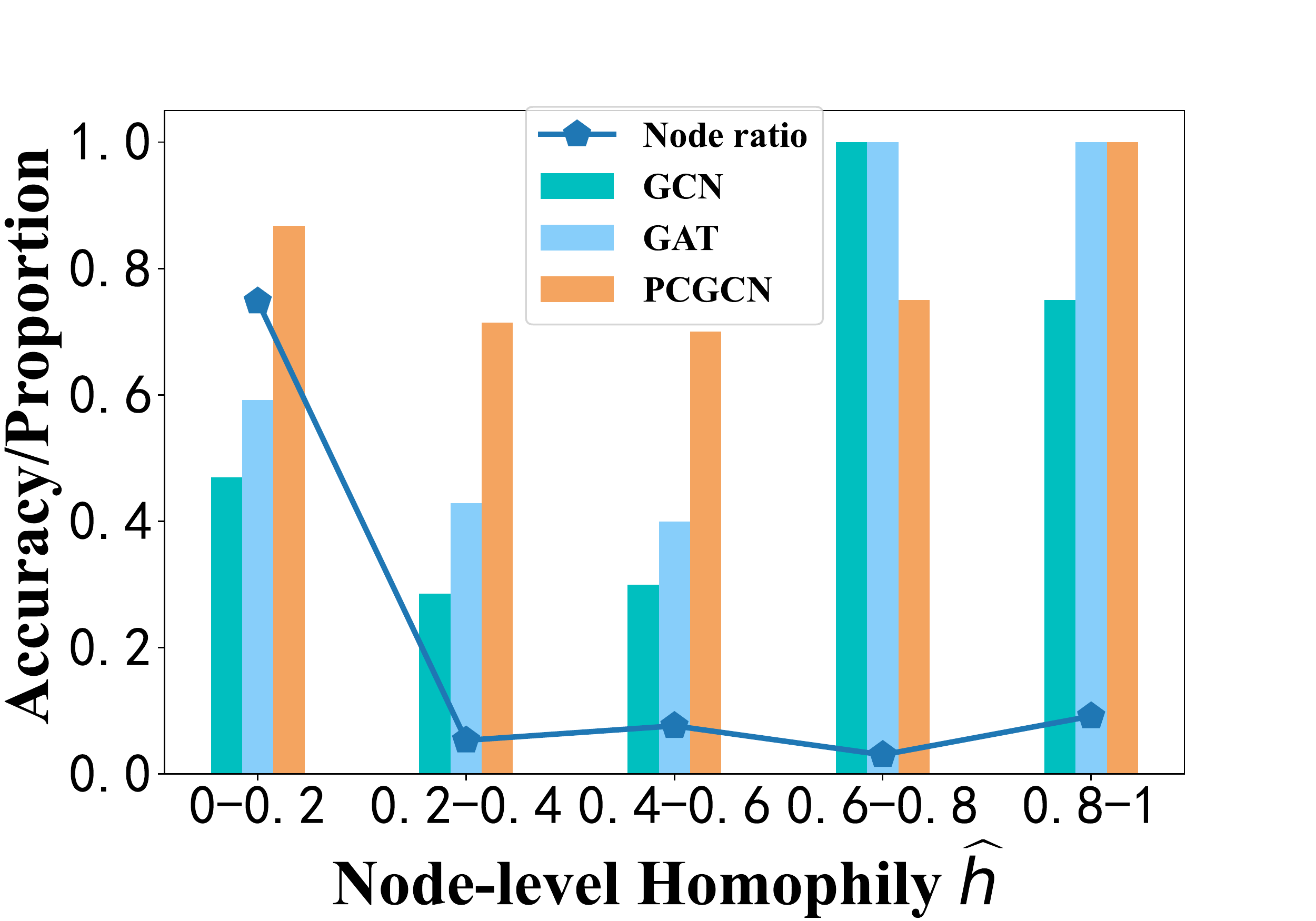}
    }
    \hspace{-0.5cm}
    \subfigure[Connell]
    {
    \includegraphics[scale=0.21]{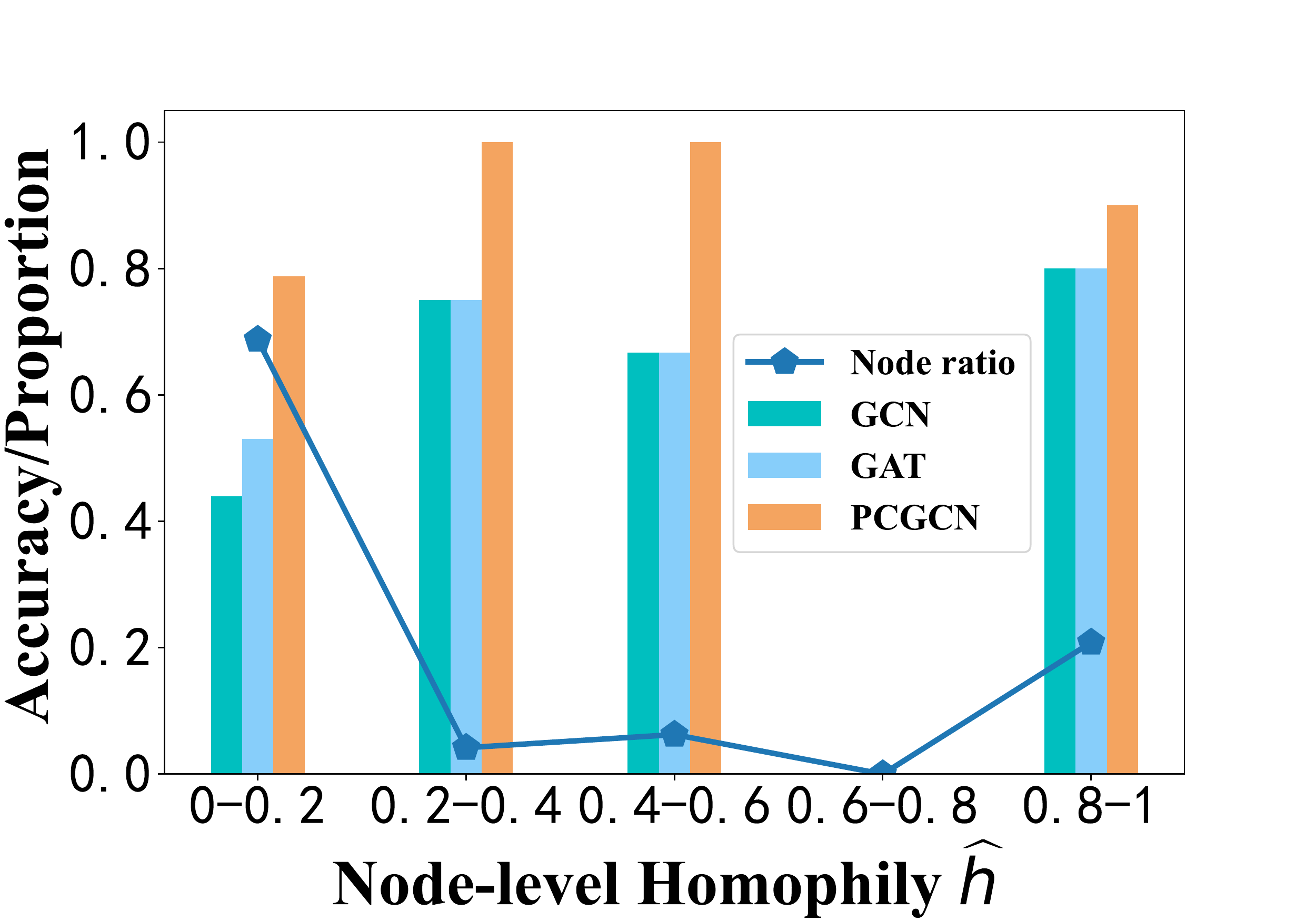}
    }
    \caption{The node classification accuracy by three models, ordered by node-level homophily ratio range. The dotted-line indicates the proportion of nodes in each homophily interval.}
    \label{fig:hete}
}
\end{figure*}

\subsubsection{Comparison with Baselines on Heterophily Issue}
To answer \textbf{Q1}, we report the test accuracy of different GNNs on the supervised node classification task over
datasets with varying homophily levels in Table~\ref{heterophily dataset}. It can be seen that PCGCN achieves the new state-of-the-art performances on almost all heterophilious graphs ($h<0.5$) with remarkable margins, compared to the best of the existing models. Moreover, PCGCN outperforms the other methods across all datasets in terms of average rank (2.0), suggesting its strong
adaptability to graphs at various homophily levels. In particular, for heterophilous datasets like Chameleon and Squirrel, PCGCN
improves the accuracy by around 3.1\% and 3.6\% compared to the second-best model, respectively. Compared with leading GNNs on homophilous graphs, PCGCN also achieves competitive accuracy. It is noteworthy that our model is a shallow model, i.e., a 2-layer GCN with pinning control, but it is still comparable or even slightly better than the deep GNN model GCNII with 64 layers. All these results demonstrate that PCGCN's hybrid message passing effectively facilitates the vanilla message passing GNNs. 

It is interesting to explore the performance of PCGCN on specific nodes with different local homophily levels. Figure~\ref{fig:hete} shows the classification accuracy of PCGCN and of two vanilla message passing GNNs, namely GCN and GAT, on the nodes with varying node-level homophily. Clearly, PCGCN is superior to the vanilla message passing GNNs for nodes with low local homophily (i.e., strongly heterophilious nodes), corresponding to the node-level homophily less than 0.4, which shows that pinning control is capable of alleviating the negative effect of heterophily on node classification. It is also clear that, for homophilous nodes, i.e., the nodes with node-level homophily greater than $0.6$, PCGCN is only comparable with GCN. Although in heterophilious graphs the majority of nodes are weakly homophilous, a significant improvement of the overall node classification performance by PCGCN is still achieved, as shown in Table~\ref{heterophily dataset}.

\begin{figure*}[htbp]
	\centering
	{
	\subfigure[initial feature]
	{
	\hspace{-0.5cm}
    \includegraphics[width=0.25\textwidth]{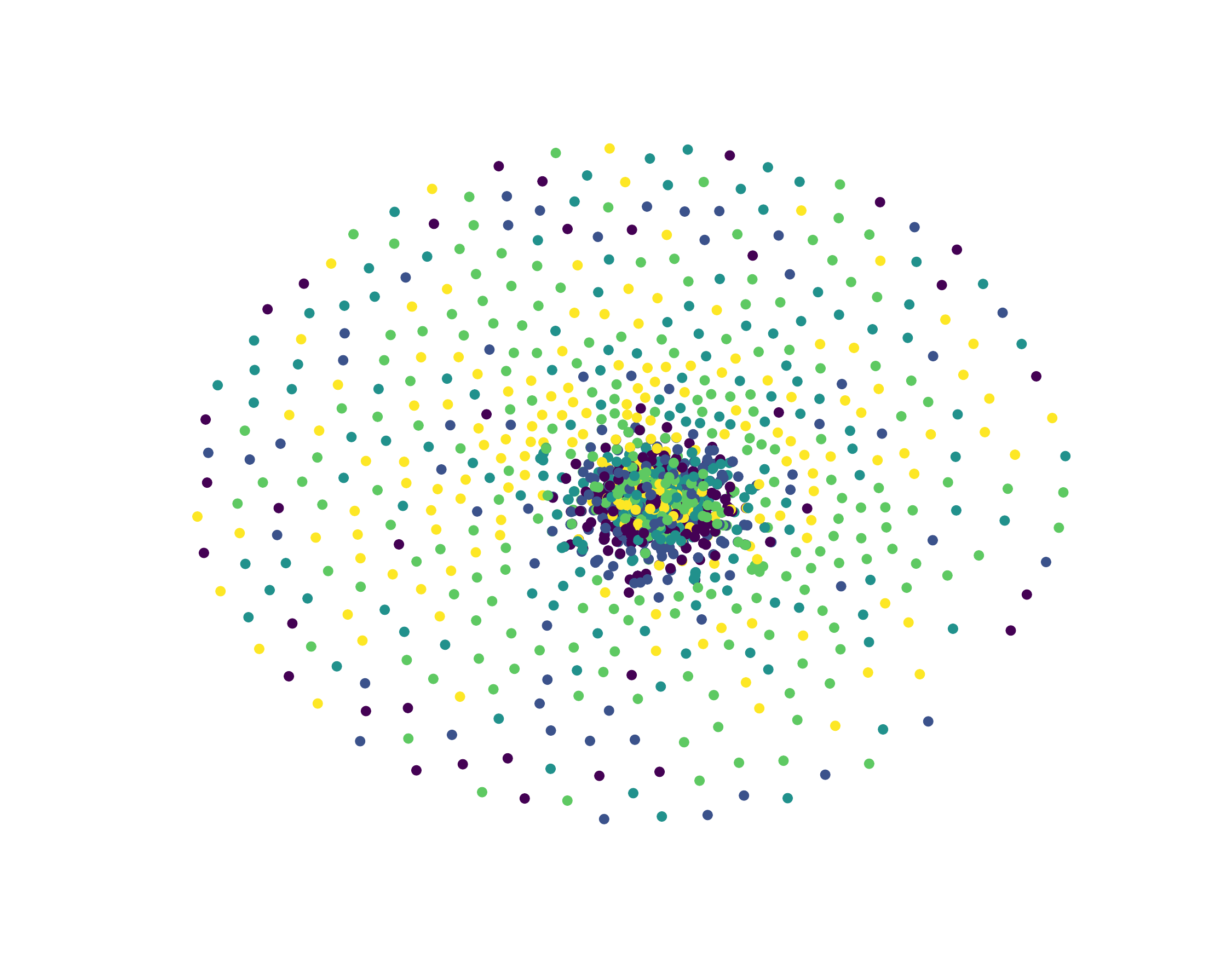}}
	\subfigure[GCN]
	{
	\hspace{-0.5cm}
    \includegraphics[width=0.25\textwidth]{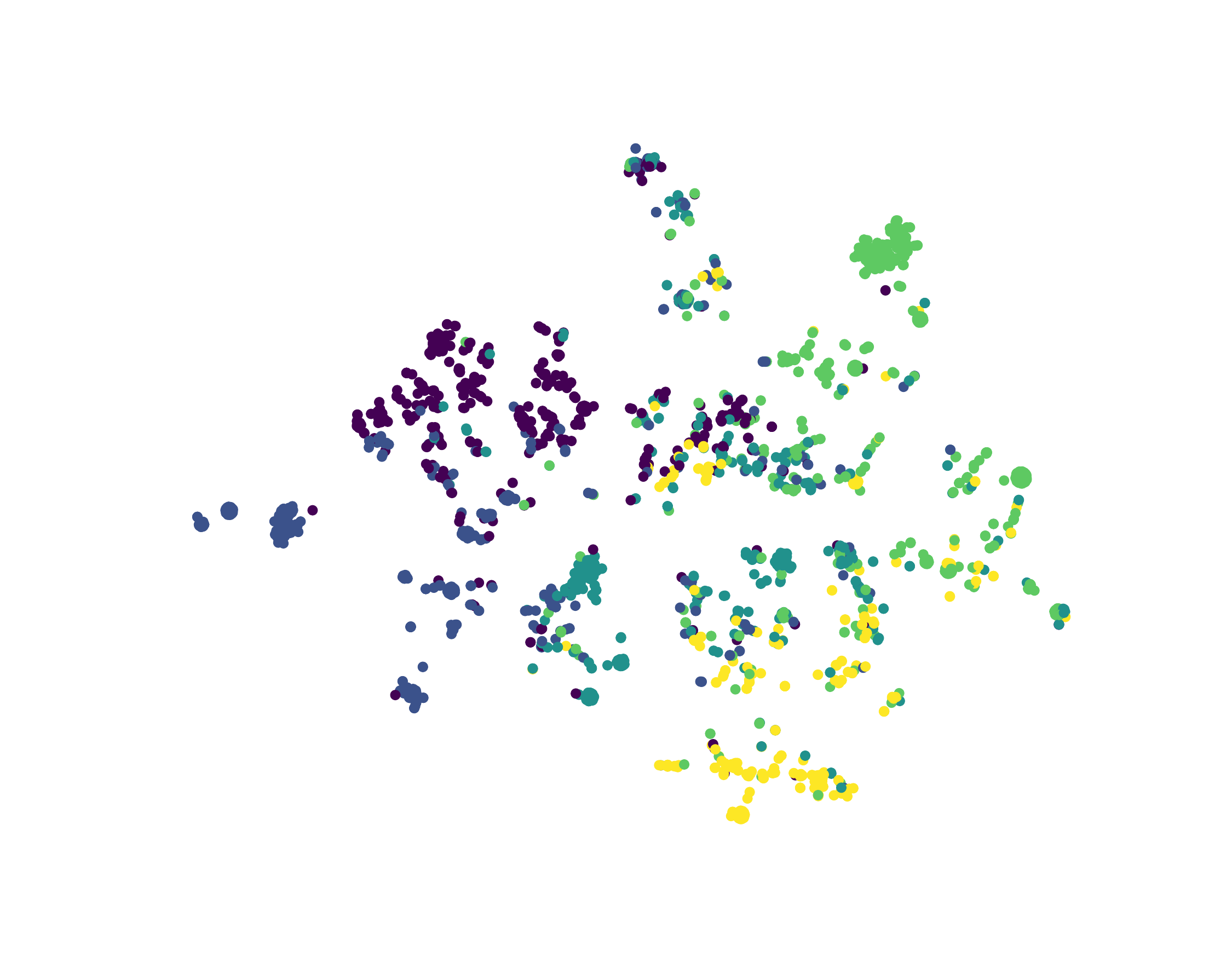}}
    \subfigure[LINKX]
	{
	\hspace{-0.5cm}
    \includegraphics[width=0.25\textwidth]{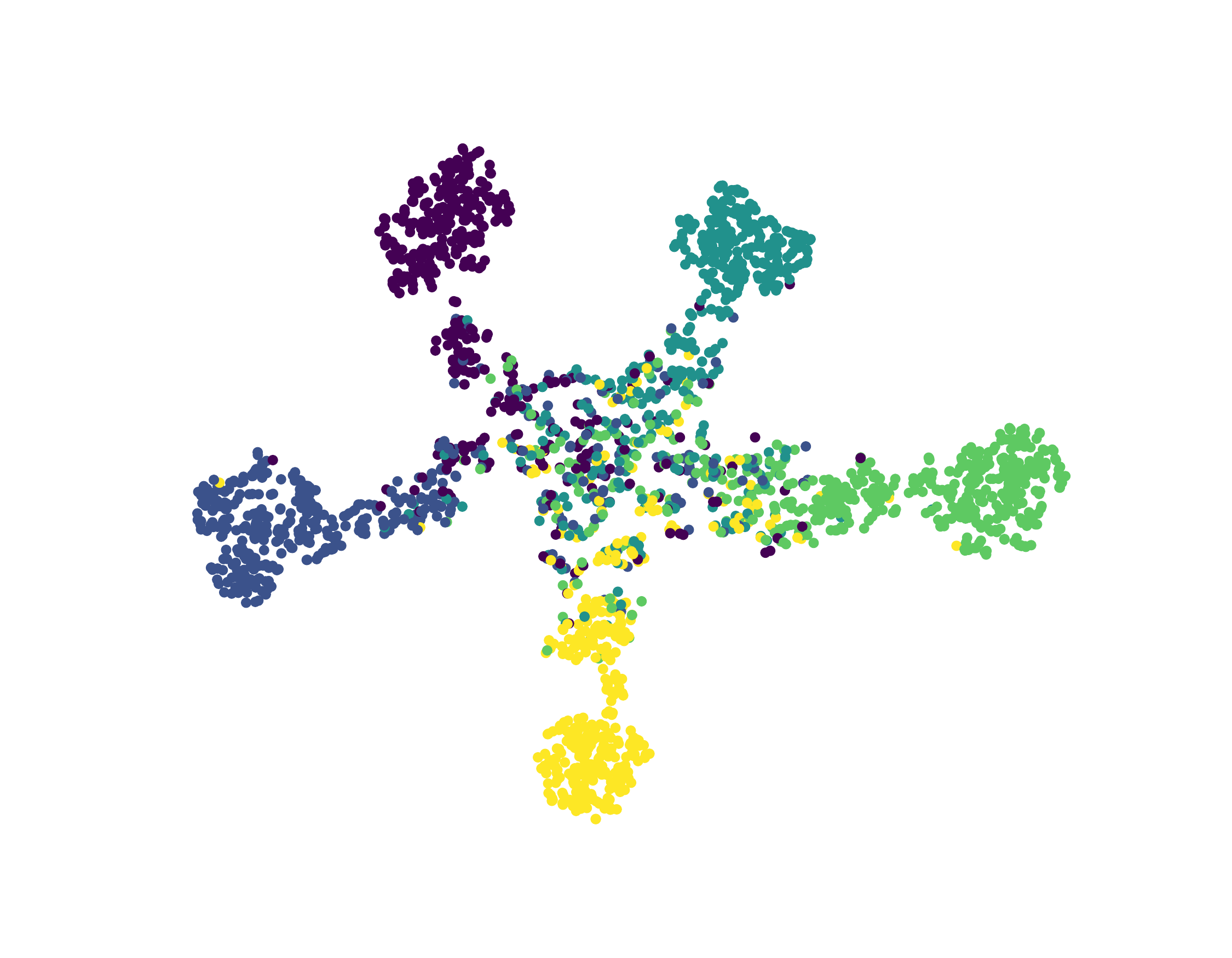}}
    \subfigure[PCGCN]
    { 
    \hspace{-0.5cm}
    \includegraphics[width=0.26\textwidth]{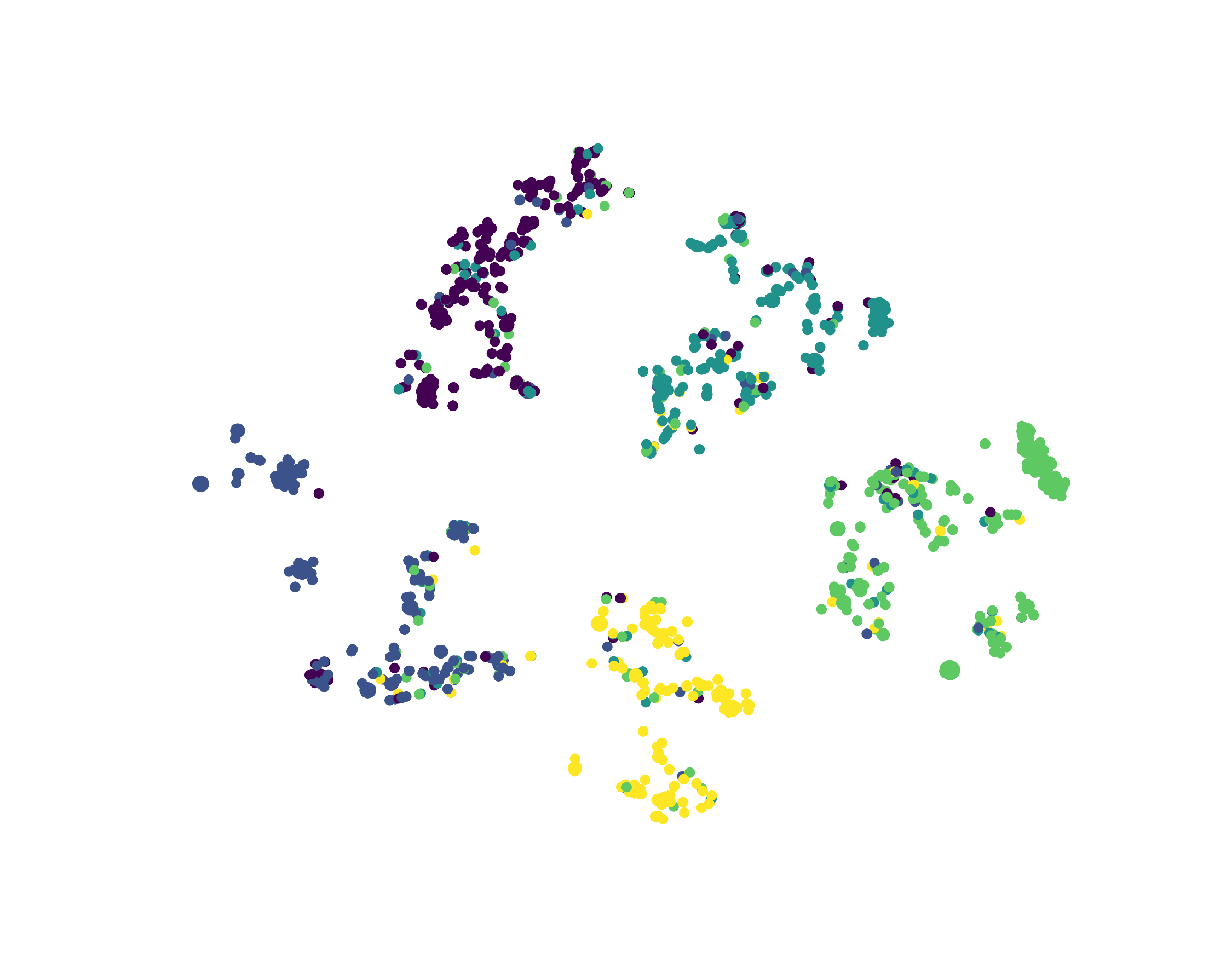}}
	\caption{t-SNE visualization of node representations learned by GCN, LINKX and PCGCN on Chameleon, respectively.}
	\label{fig:tsne1}
}
\end{figure*}

\begin{figure*}[htbp]
	\centering
	{
    \subfigure[initial feature]
	{
	\hspace{-0.5cm}
    \includegraphics[width=0.25\textwidth]{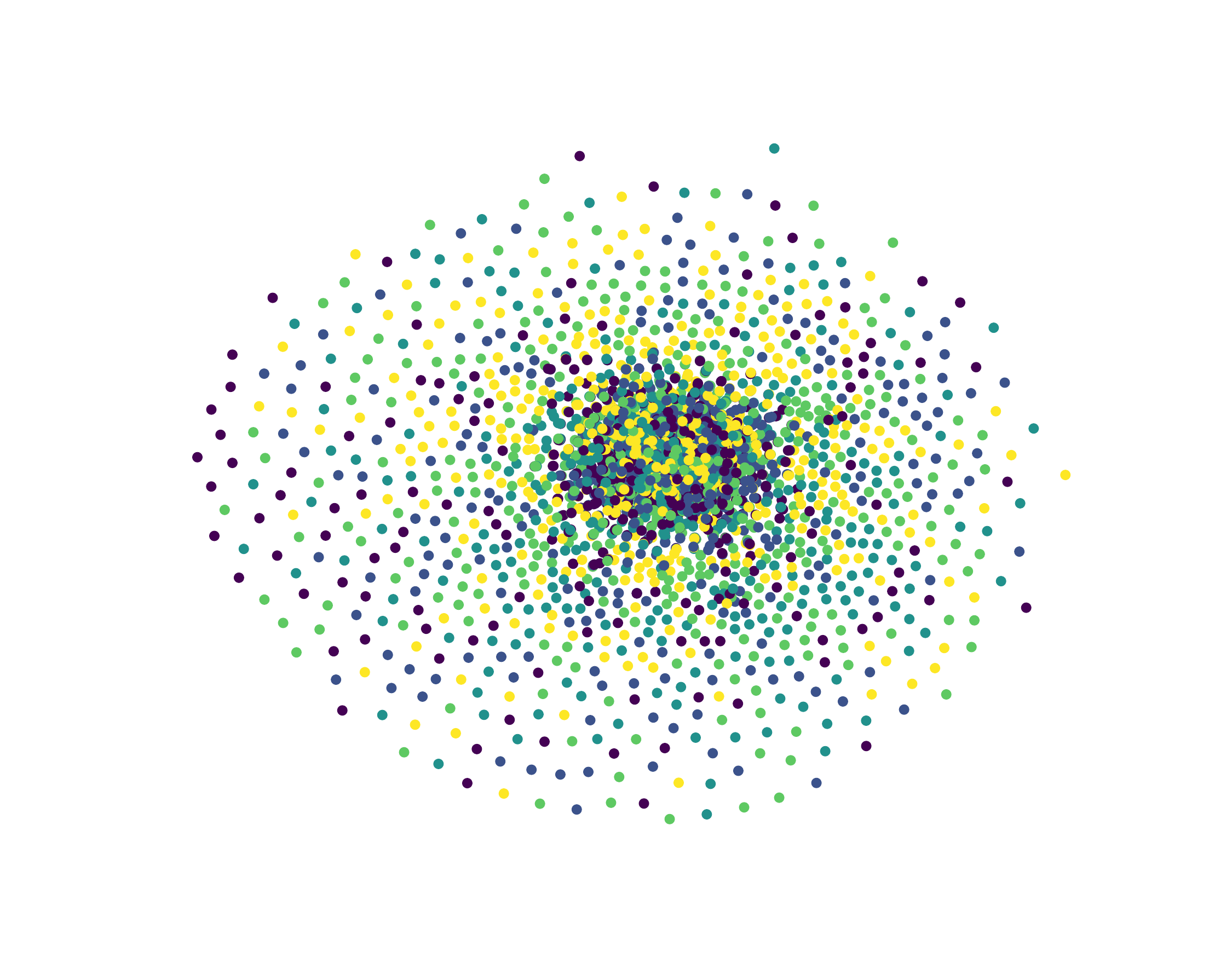}}
    \subfigure[GCN]
	{
	\hspace{-0.5cm}
    \includegraphics[width=0.25\textwidth]{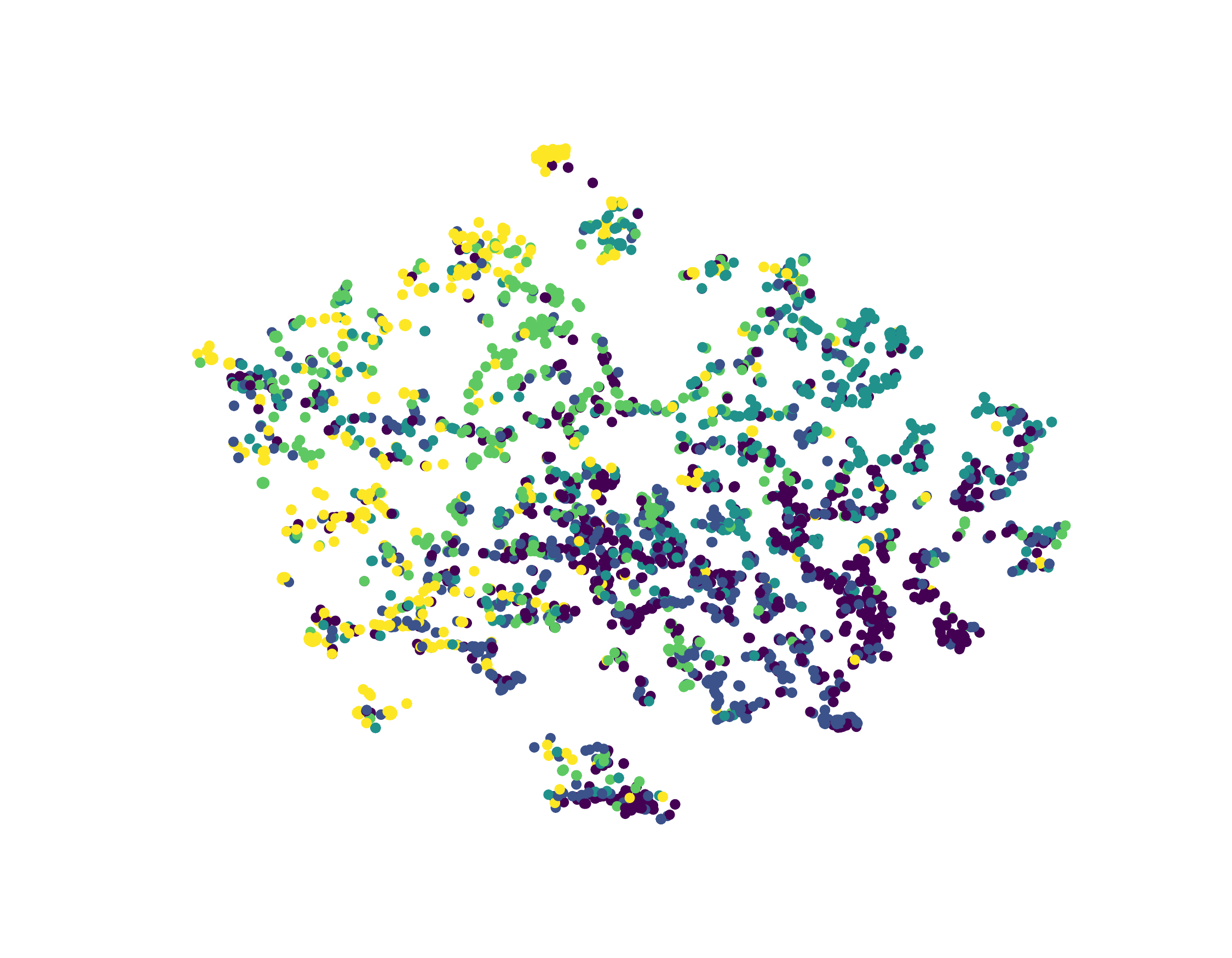}}
    \subfigure[LINKX]
	{
	\hspace{-0.5cm}
    \includegraphics[width=0.25\textwidth]{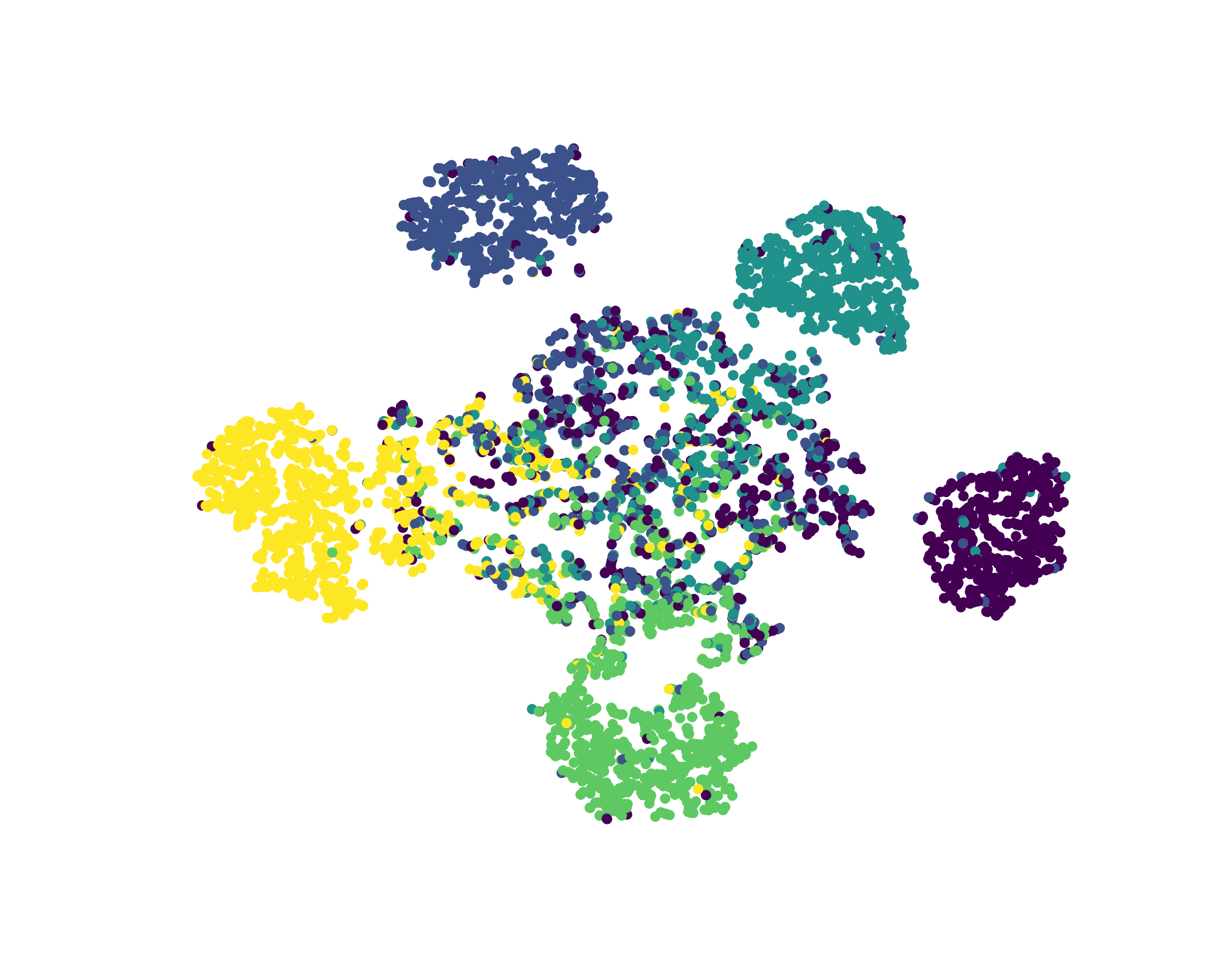}}
    \subfigure[PCGCN]
    { 
    \hspace{-0.5cm}
    \includegraphics[width=0.25\textwidth]{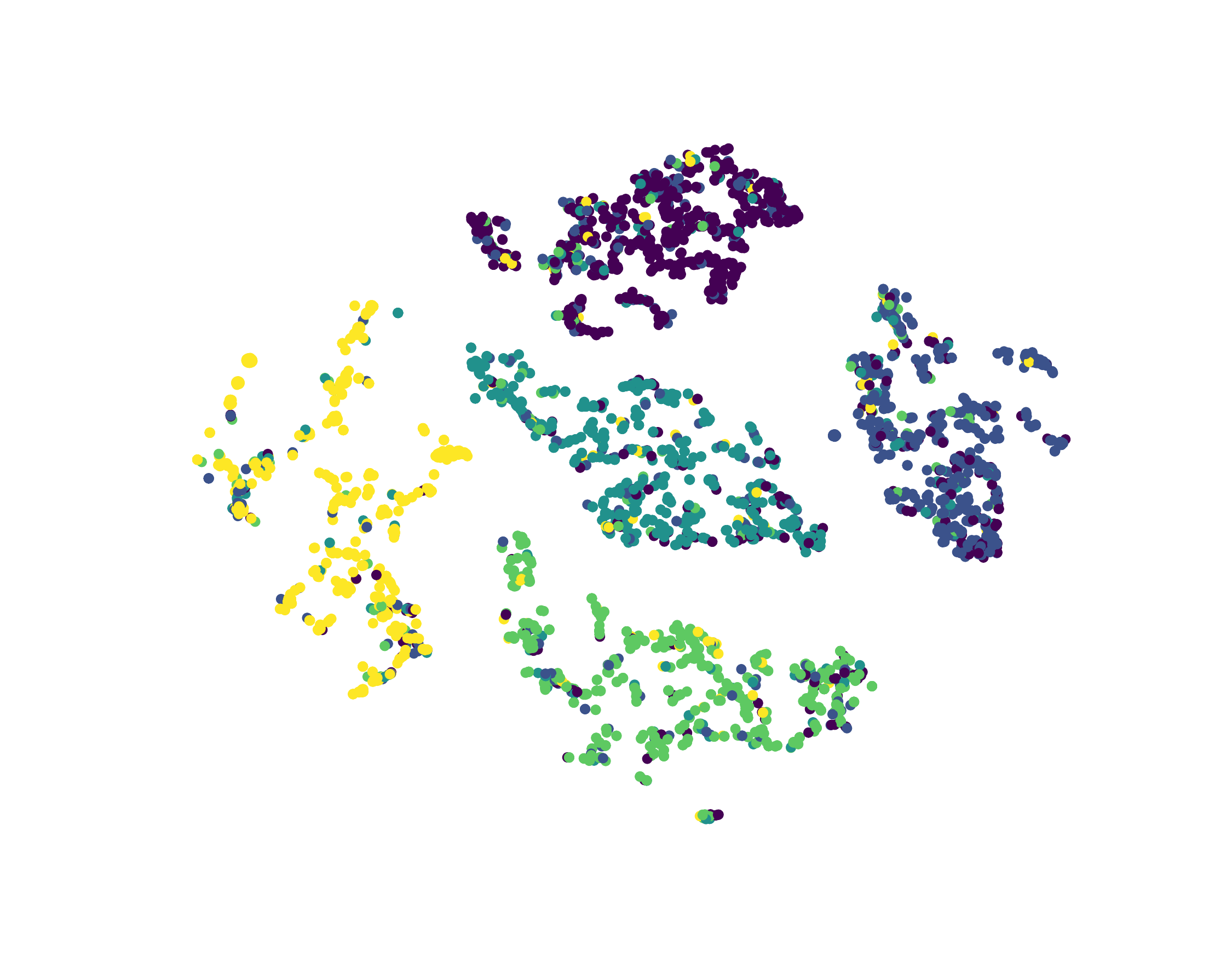}}
	\caption{t-SNE visualization of node representations learned by GCN, LINKX and PCGCN on Squirrel, respectively.}
	\label{fig:tsne2}
}
\end{figure*}

To intuitively understand what change our pinning control brings to node representation learning, we visualize node feature distribution in the embedding space. We utilize t-SNE to create 2D plots of all node embeddings at the last layer after training for two heterophilious graphs: chameleon and squirrel. Figure~\ref{fig:tsne1} and Figure~\ref{fig:tsne2} show the node embedding distributions of two datasets achieved by GCN, LINKX and PCGCN, respectively, where different colors indicate different classes. It is clear that, compared to the random distribution of initial node features, some clustering patterns in the feature subspace are captured by GCN but lack of obvious boundary. While LINKX produces a clearer clustering structure than GCN, there is still a large portion of overlap between classes. In contrast, PCGCN with adaptive parameter learning (corresponding to the last subplot in each row) assigns nodes into several distinct clusters, significantly reducing the representation noise thereby improving the classification performance.

\begin{table*}[htbp]
	\centering
	\renewcommand\arraystretch{1.3}
	\setlength\tabcolsep{7.0pt}
	\caption{Results with different missing classes on datasets in terms of classification accuracy (in percentage).}
	\label{tb:missclass-1}
	\begin{tabular}{lc|cccccc}
		\hline
		\textbf{Dataset} & \textbf{Chameleon}& \textbf{Chameleon-1} & \textbf{Chameleon-2} & \textbf{Chameleon-3} & \textbf{Chameleon-4} & \textbf{Chameleon-5} & \textbf{Average}\\
		\hline
		MLP & 46.93±1.7& 23.68±3.9	&23.83±4.4	&26.62±6.9	&21.86±3.3	&26.09±6.7 & 24.41
\\
		GCN &65.92±2.5 &38.35±4.0	&40.54±5.0	&37.82±4.2	&34.73±4.4	&36.86±2.8 & 37.65
\\
		GAT& 65.32±1.9 &35.83±4.2	&36.60±2.5&	40.50±6.9&	33.88±2.8&	37.10±2.8 & 36.78
 \\
		PCGCN & \textbf{74.29±1.9}& \textbf{57.71±1.9}	& \textbf{58.09±2.0} &\textbf{61.62±1.8}	&\textbf{56.35±1.7}	&\textbf{56.11±2.6}&\textbf{57.96}
\\
		\hline
	\end{tabular} 
\end{table*}

\begin{table*}[htbp]
	\centering
	\renewcommand\arraystretch{1.3}
	\setlength\tabcolsep{10.2pt}
	\caption{Results with different missing classes on datasets in terms of classification accuracy (in percentage).}
	\label{tb:missclass-2}
	\begin{tabular}{lc|cccccc}
		\hline
		\textbf{Dataset} &\multicolumn{1}{l|}{\textbf{Squirrel}} & \textbf{Squirrel-1} & \textbf{Squirrel-2} & \textbf{Squirrel-3} & \textbf{Squirrel-4} & \textbf{Squirrel-5} & \textbf{Average}\\
		\hline
		MLP & 29.95±1.6&19.79±0.9	&20.62±1.3	&19.97±1.0	& 19.96±0.7	&20.0±0.9 & 20.06\\
		GCN& 49.78±2.0&22.35±2.6	&21.59±1.9	&21.72±2.3	&21.26±1.6	&21.87±1.4 & 21.75\\
		GAT&46.79±2.0&22.89±1.7&	22.89±1.7&	21.88±1.8&	21.73±2.1&	21.52±2.4 & 22.02\\
		PCGCN & \textbf{65.47±2.4}& \textbf{50.99±2.1} & \textbf{53.54±1.4}	& \textbf{51.39±1.2}	& \textbf{48.43±1.2} & \textbf{45.82±1.7} & \textbf{50.03}

\\
		\hline
	\end{tabular} 
\end{table*}

\begin{figure}[htbp]
	\centering
    \subfigure[Cora]
    { 
    \includegraphics[scale=0.19]{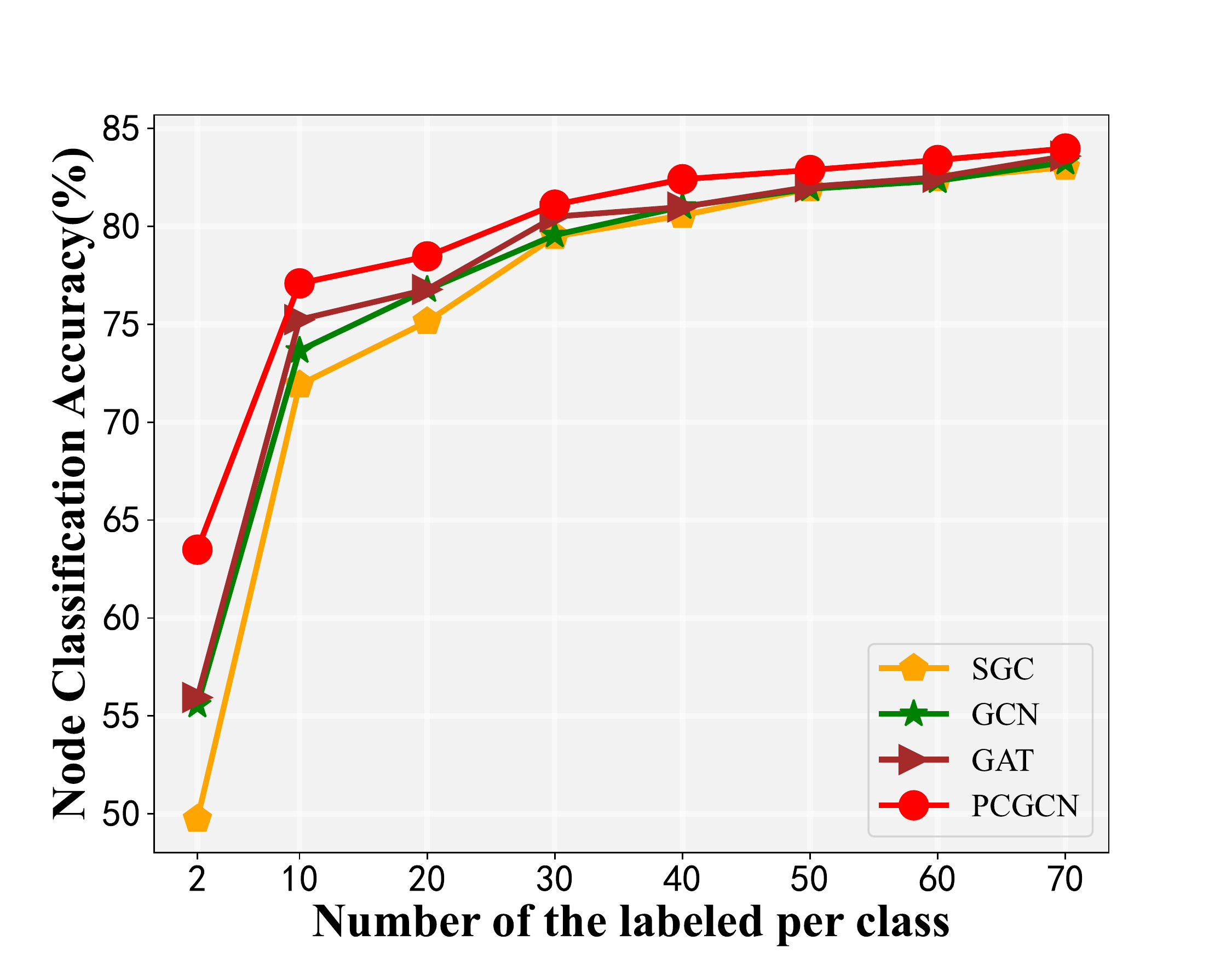}}
	\hspace{-0.5cm}
	\subfigure[CiteSeer]
	{
    \includegraphics[scale=0.19]{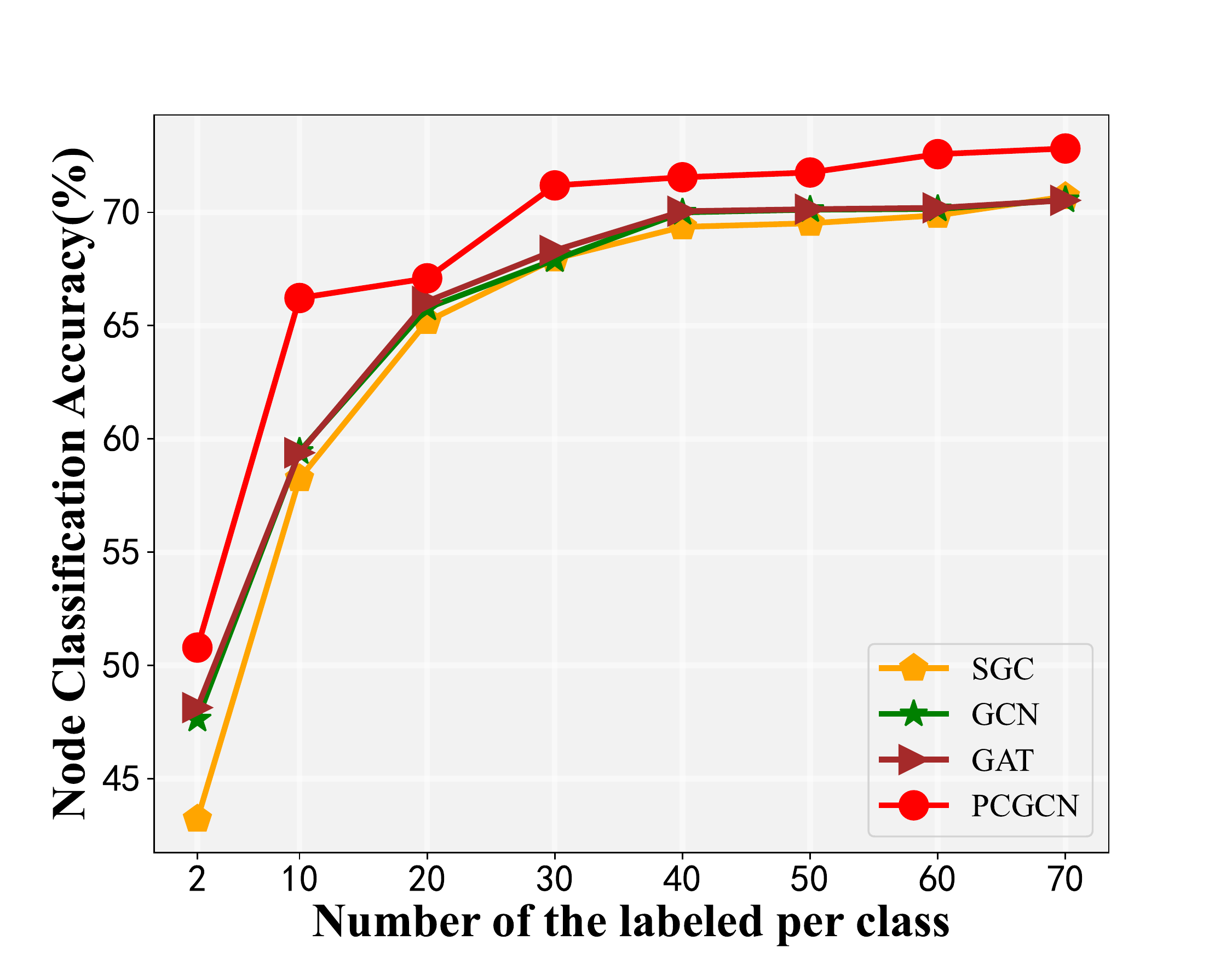}}
    \hspace{-0.5cm}
    \subfigure[Pubmed]
    { 
    \includegraphics[scale=0.19]{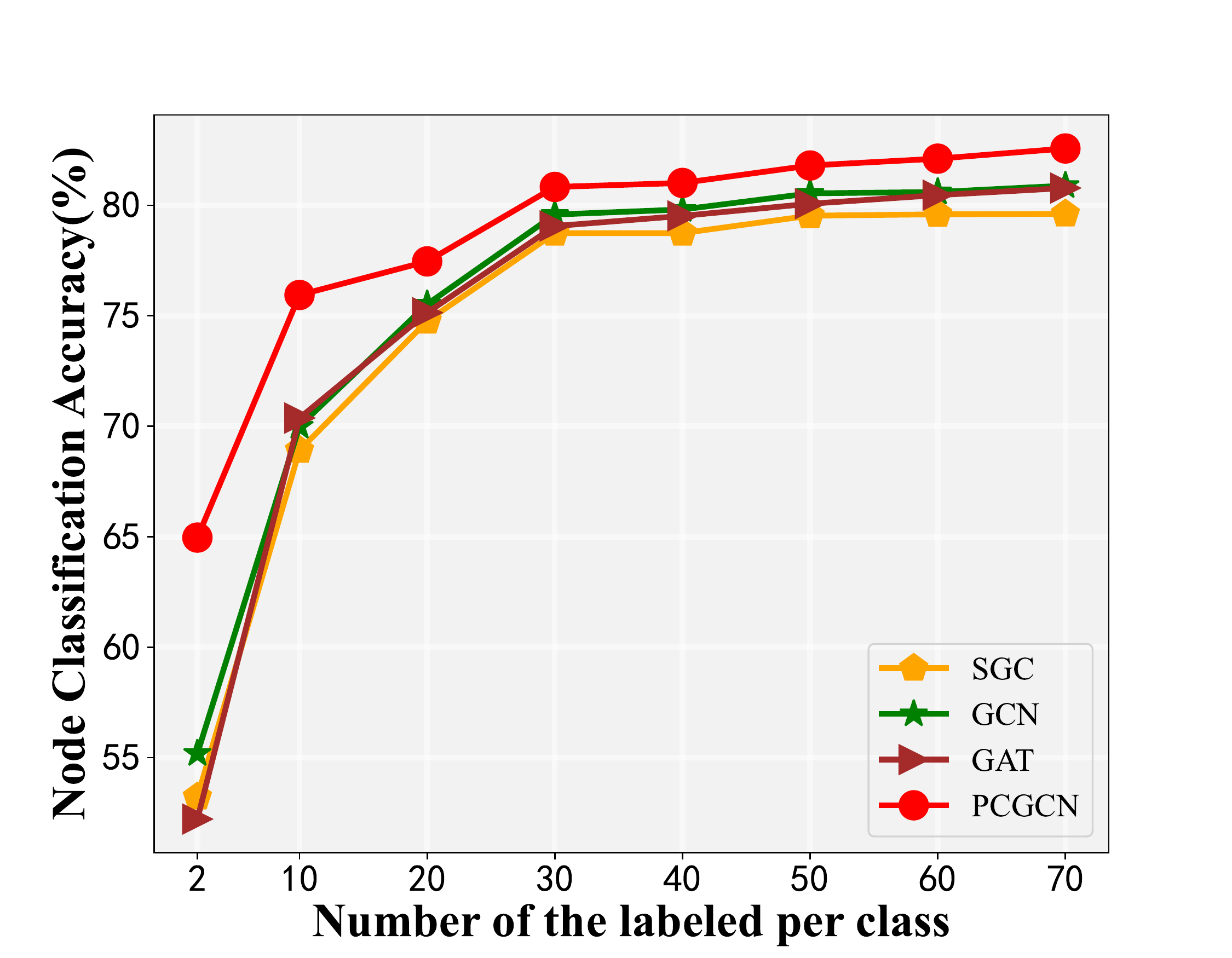}}
    \hspace{-0.5cm}
	\subfigure[Chameleon]
	{
    \includegraphics[scale=0.19]{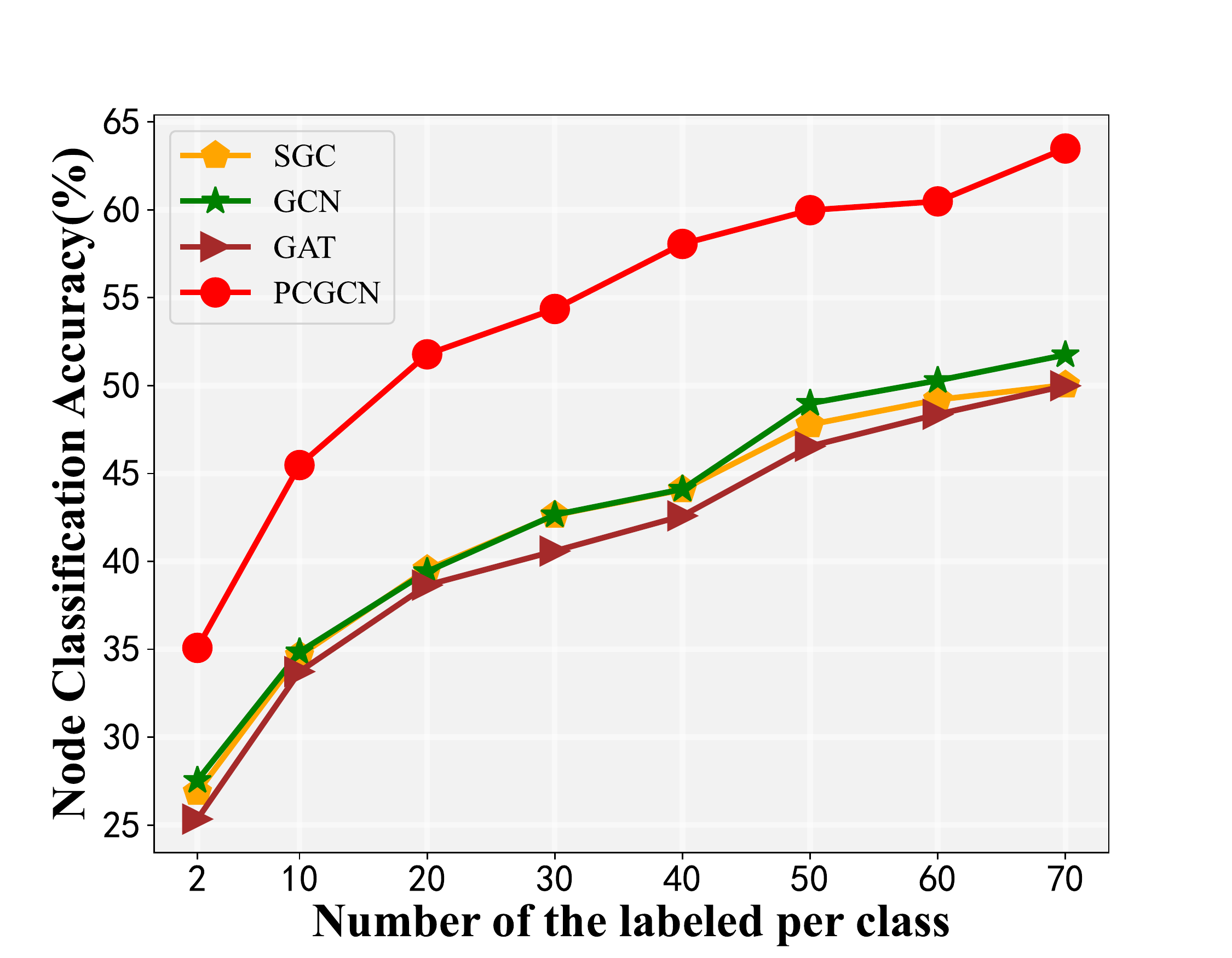}}
    \hspace{-0.5cm}
	\subfigure[Squirrel]
	{
    \includegraphics[scale=0.19]{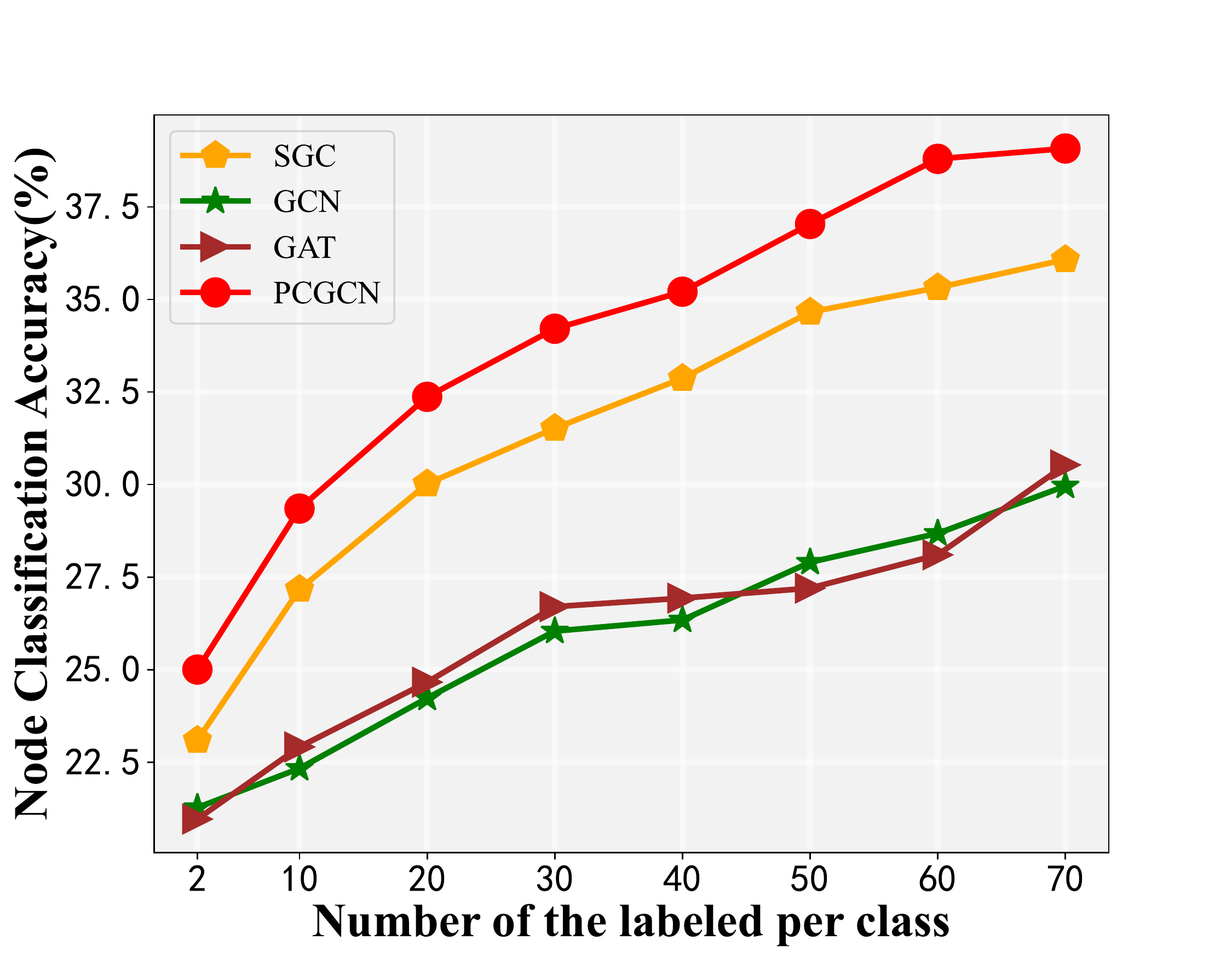}}
    \hspace{-0.5cm}
    \subfigure[Actor]
	{
    \includegraphics[scale=0.19]{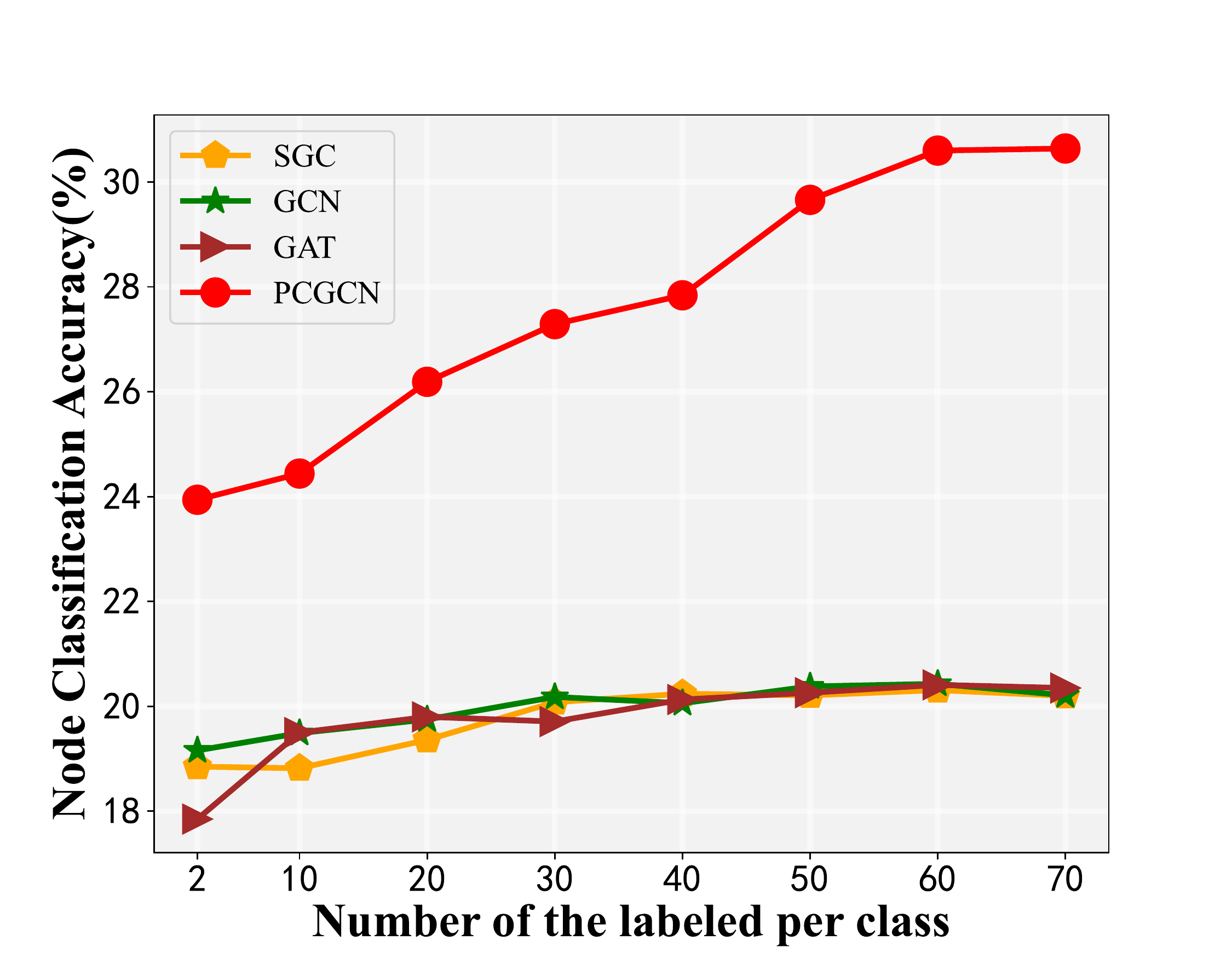}}
	\caption{Performance comparison with small numbers of labels on Homophilous and Heterophilious datasets.}
	\label{fig:label}
\end{figure}

\begin{figure}[!t]
	\centering
	\subfigure[Cora($\widetilde{h}=0.81$)]
	{
    \includegraphics[width=0.23\textwidth]{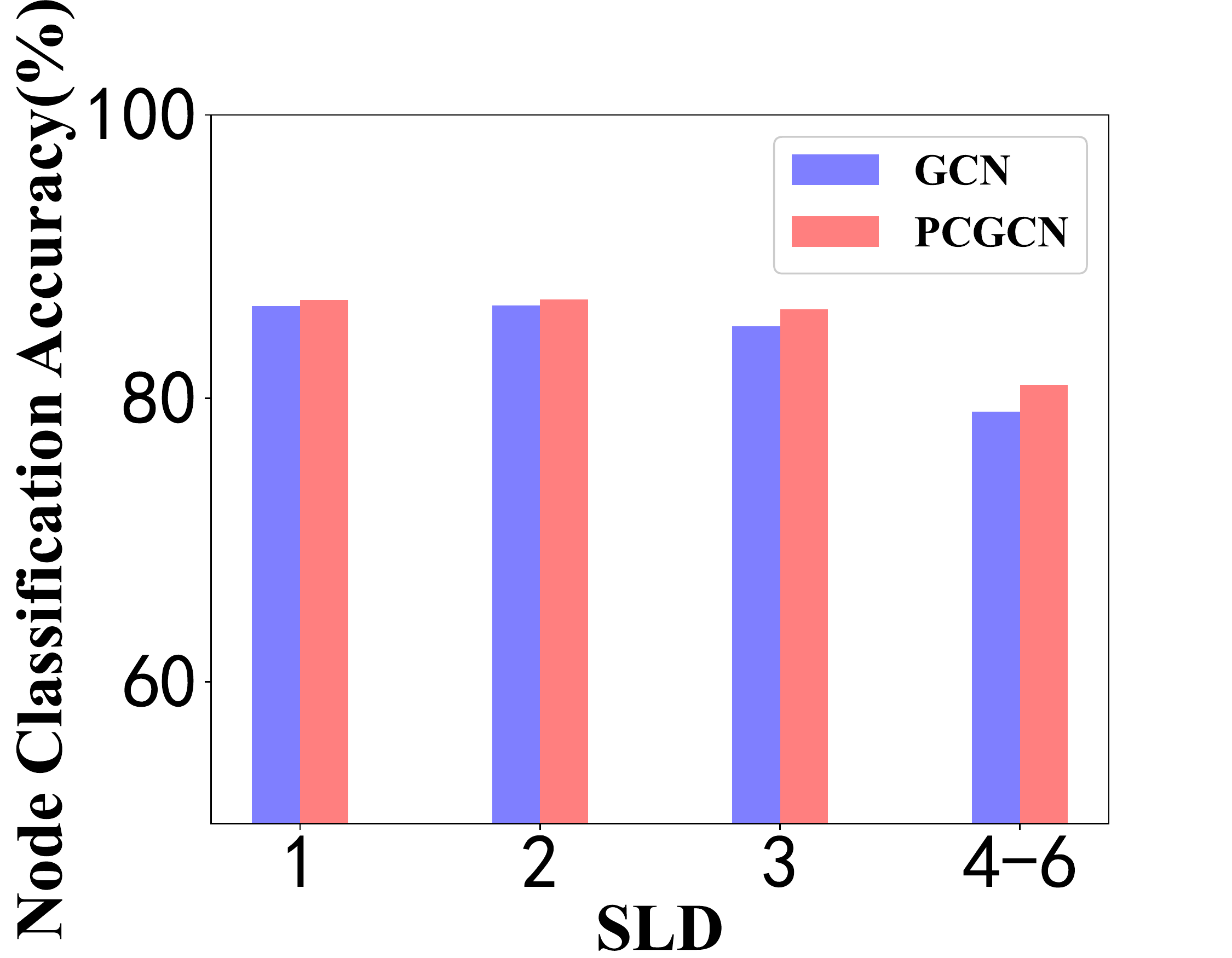}}
    \hspace{-0.5cm}
    \subfigure[CiteSeer($\widetilde{h}=0.74$)]
    {
    \includegraphics[width=0.23\textwidth]{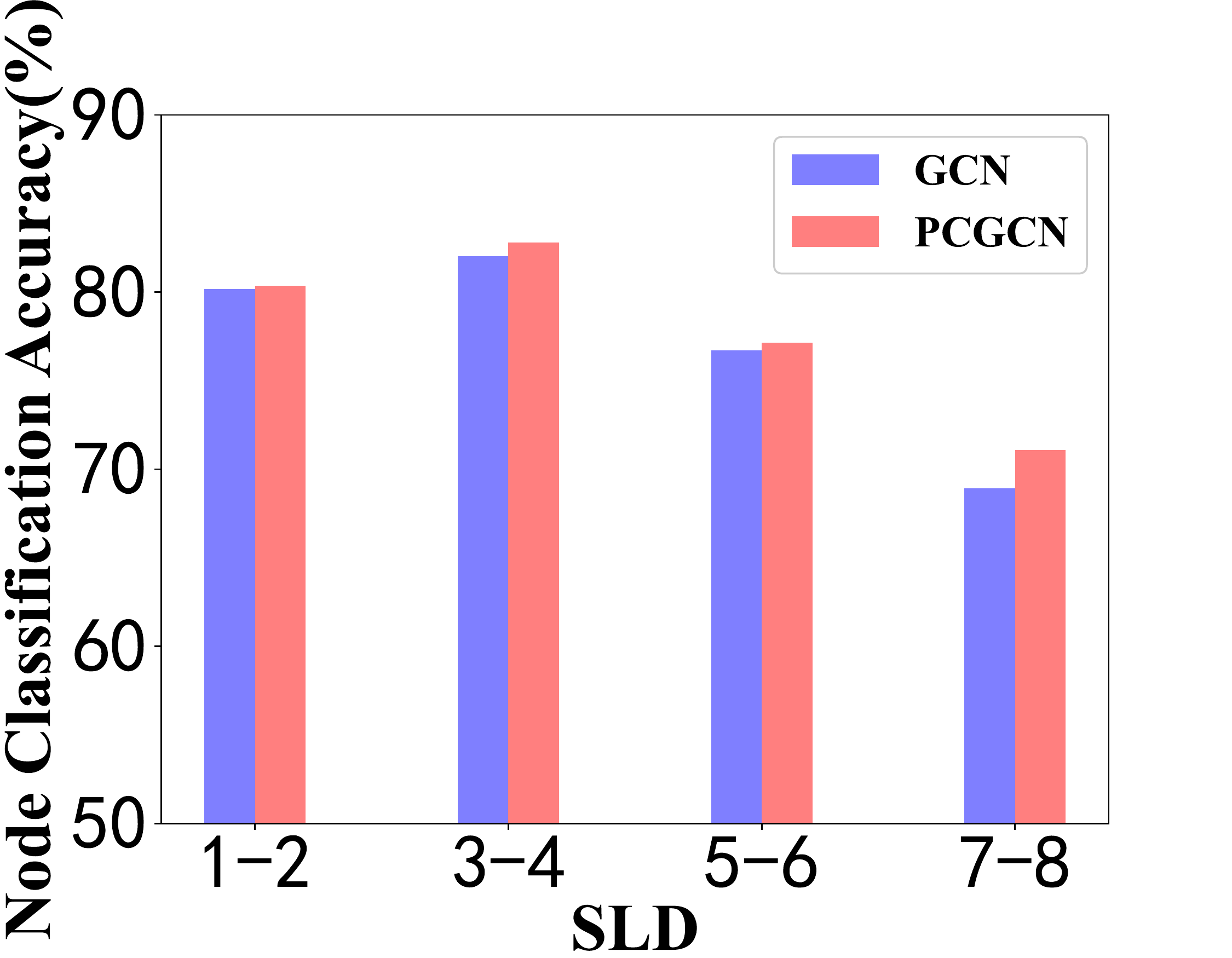}}
    \hspace{-0.5cm}
    \subfigure[Pubmed($\widetilde{h}=0.8$)]
    {
    \includegraphics[width=0.23\textwidth]{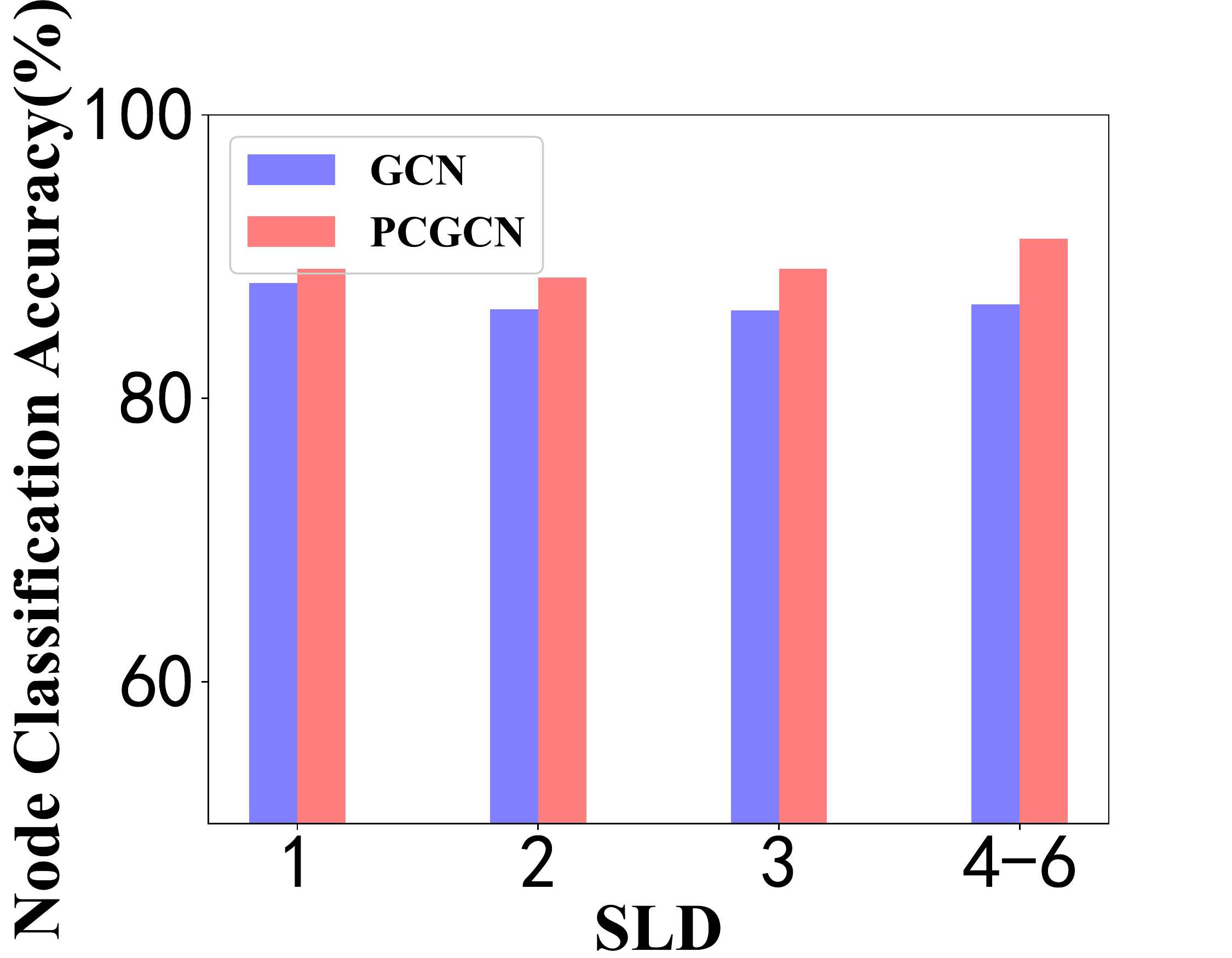}}
    \hspace{-0.5cm}
    \subfigure[Chameleon($\widetilde{h}=0.23$)]
    { 
    \includegraphics[width=0.23\textwidth]{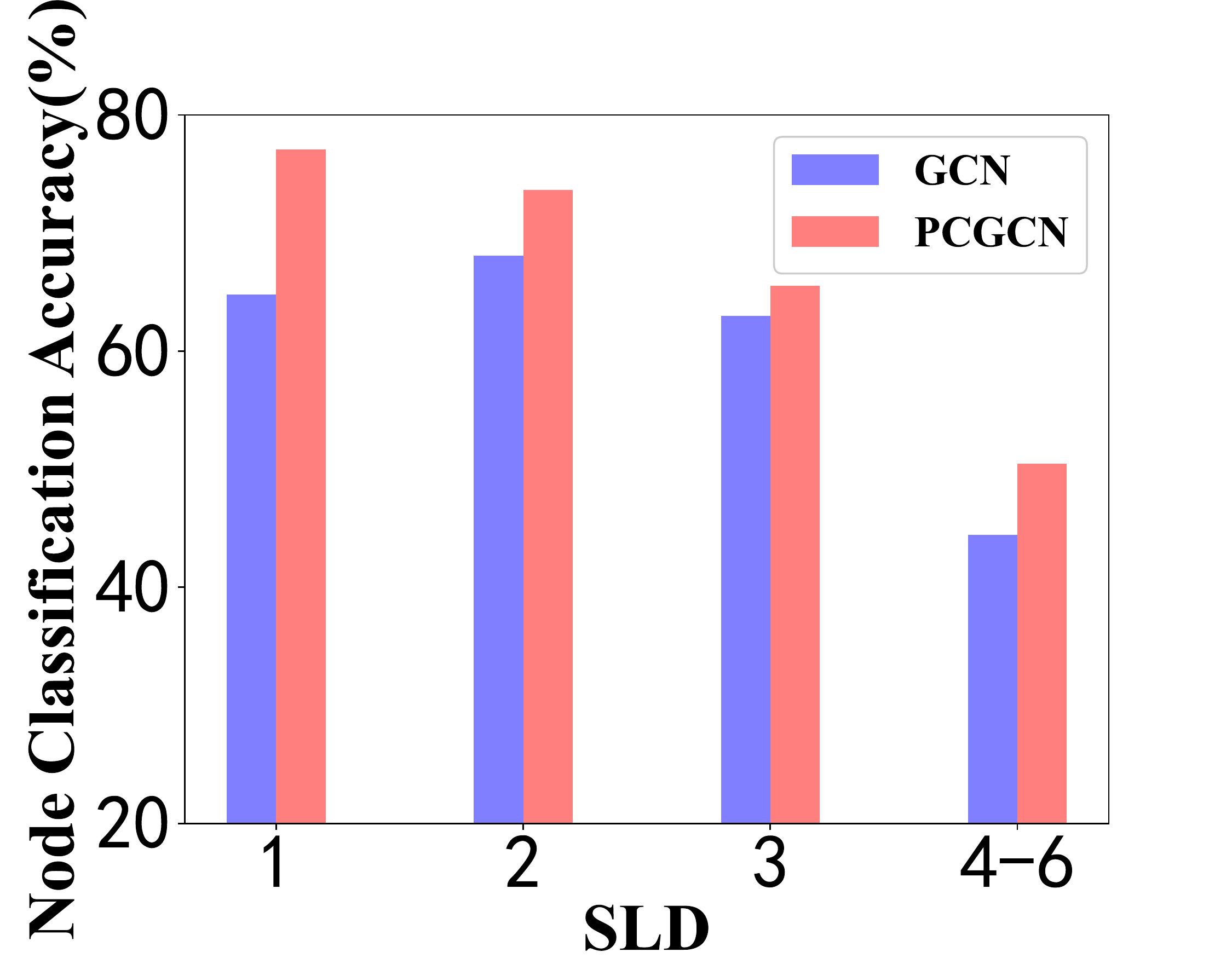}}
    \hspace{-0.5cm}
    \subfigure[Squirrel($\widetilde{h}=0.22$)]
    { 
    \includegraphics[width=0.23\textwidth]{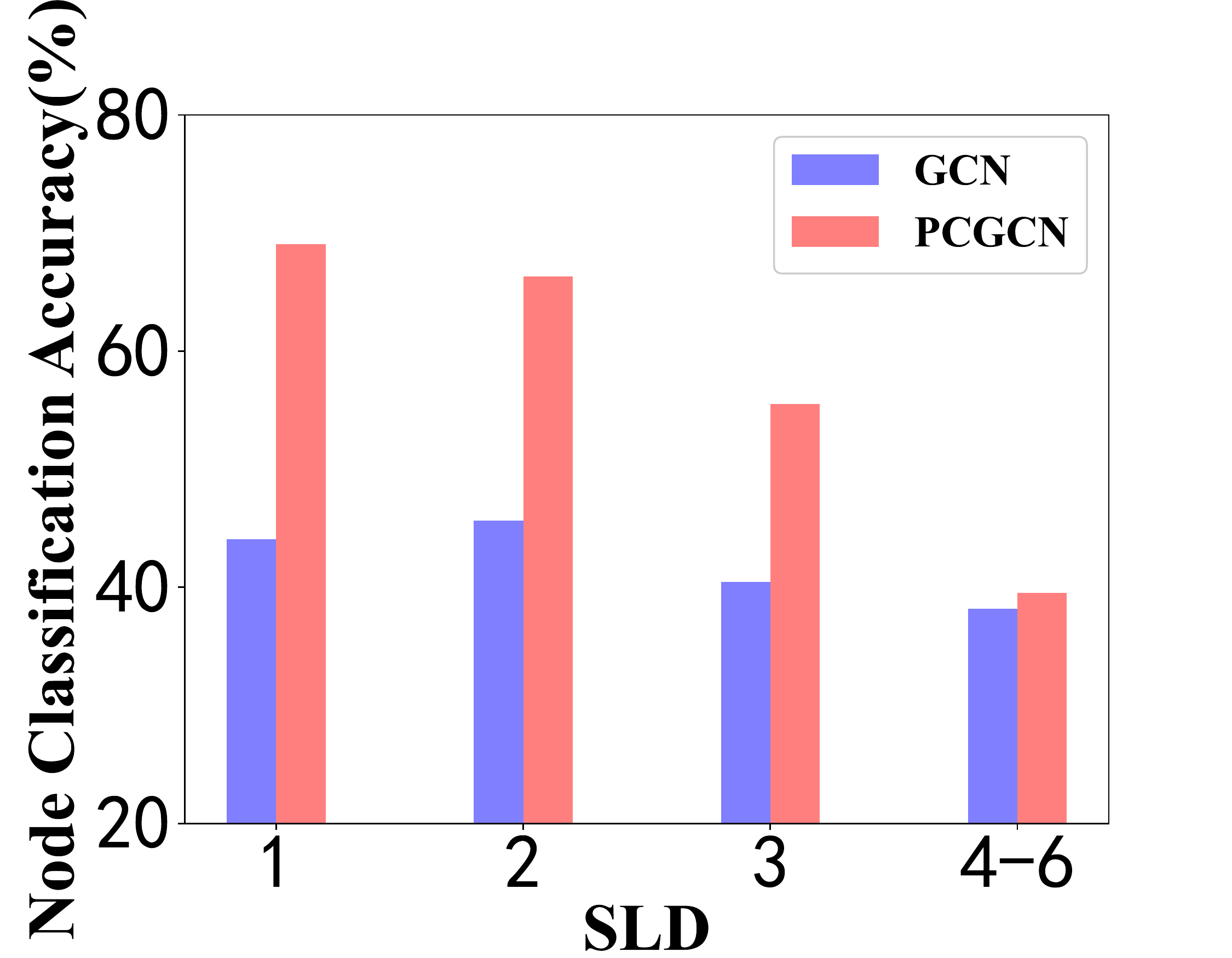}}
    \hspace{-0.5cm}
    \subfigure[Actor($\widetilde{h}=0.22$)]
    { 
    \includegraphics[width=0.23\textwidth]{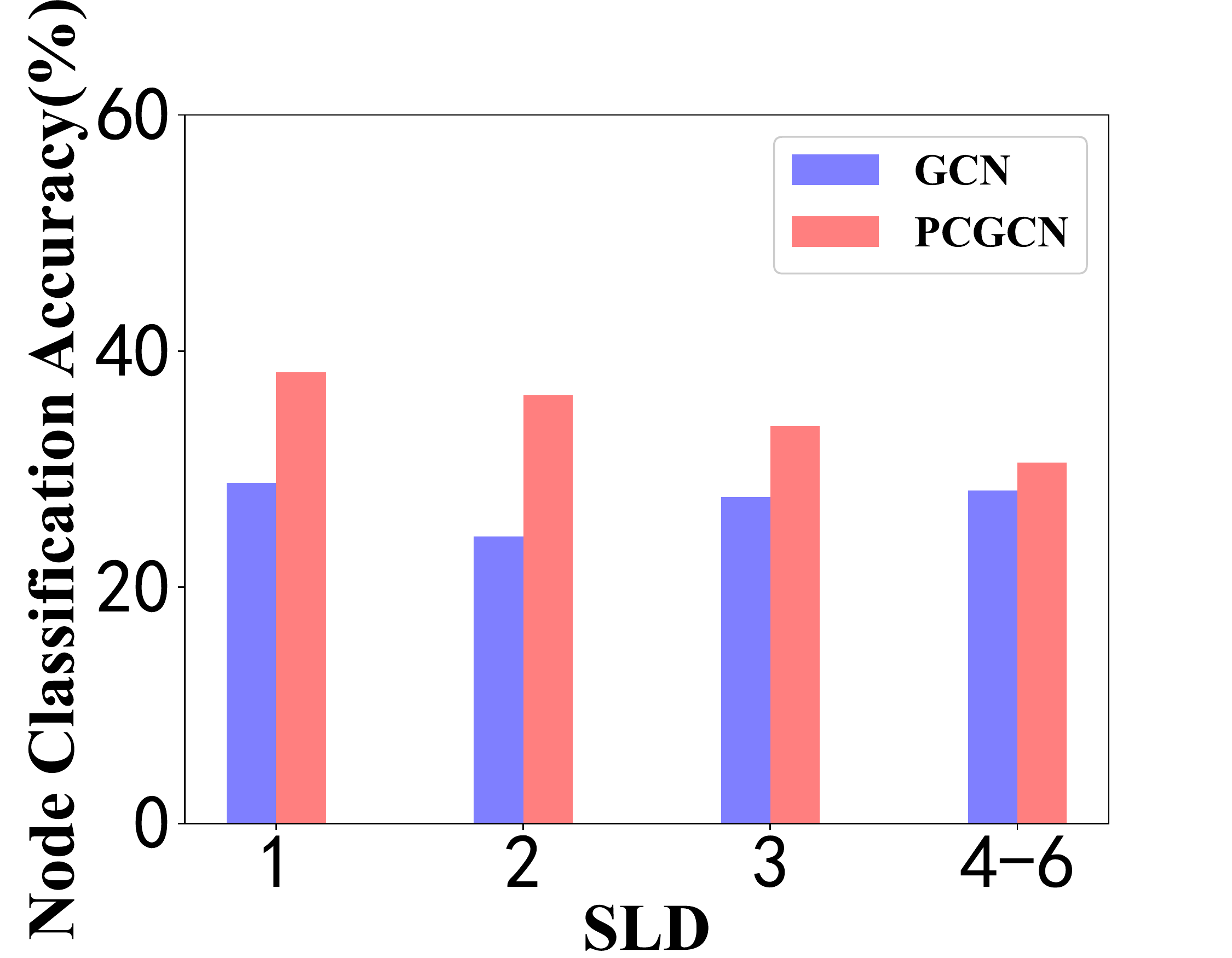}}
	\caption{Performance distribution v.s. SLD on six datasets. }
	\label{fig:SLD}
\end{figure}

\begin{figure}[!t]
\centering
\includegraphics[width=2.5in]{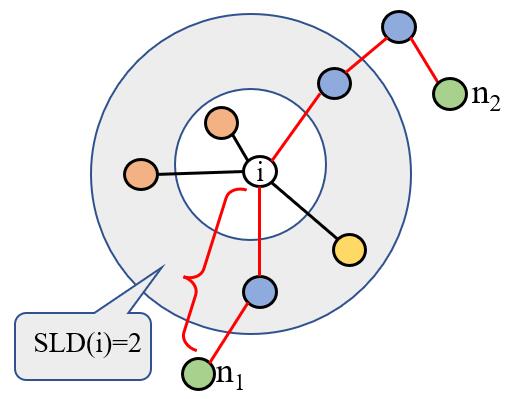}
\caption{A toy example for illustrating the calculation of SLD. Different colors represent different class labels. The ground-truth label of node $i$ is “green”, and the shortest distance from node $i$ to the “green” nodes is here $2$-hops.}
\label{fig:SLDs}
\end{figure}
\subsubsection{Efficacy of Pinning Control on Distant Nodes}
We investigate the impact of pinning control on distant nodes, that is, whether pinning control can enhance message passing for the nodes not directly connected to the labeled nodes. Towards this end, we measure the performance of PCGCN on the unlabeled nodes with varying shortest distances from the label nodes in the same class, which is graphically illustrated in Figure~\ref{fig:SLDs} and calculated as follows:
let $\mathcal{N}(1)=A$ be the one-hop neighboring matrix. The $k$-hop neighboring matrix is then obtained by performing $k$ iterations as follows:
 \begin{equation}\label{eq:neibor}
	\mathcal{N}(k)=\begin{cases}
k, & if~A_{ij}^{k}>0 ~\&~  \mathcal{N}_{ij}(k-1)=0\\ 
\mathcal{N}(k-1), & otherwise
\end{cases}
\end{equation}
%\end{myDef}
In general, $k$ can be set to be equal to or greater than the diameter of graph $G$, to guarantee that all nodes have been visited at least once, i.e., $\mathcal{N}(k) > 0$. Clearly, the elements of $\mathcal{N}(k)$ are the shortest path-lengths between nodes. Then, the shortest label distance for an arbitrary unlabeled node $i$ of class $C_i$ is the smallest value among all the shortest path-lengths between node $i$ and the labeled nodes belonging to the same class $C_i$, which formally reads as
\begin{equation}
	SLD(i) = min\begin{Bmatrix}
\mathcal{N}_{ij}(k), j\in \boldsymbol{T}(C=C_i)
\end{Bmatrix},
\end{equation}
where $\boldsymbol{T}(C=C_i)$ refers to the labeled nodes with class label $C_i$ in the training set $T$. 

%we first investigate whether pinning control affects the messages passing from randomly distributed labeled nodes to unlabeled ones. We define an metric to quantify the distance between an unlabeled node to the nearest labeled node in its class, termed shortest label distance (SLD for short). We include the definition of SLD in Appendix B. 

Figure~\ref{fig:SLD} reports the node classification accuracy for different types of test nodes in terms of SLD values. One observation is that, on homophilous graphs (i.e., Cora, CiteSeer and Pubmed), pinning control can improve the accuracy of node classification for the nodes that are far away from the labeled nodes in the same class (corresponding to large SLD values), compared to the vanilla message passing, suggesting that the pinning controllers are able to transmit the information about classes to unlabeled nodes directly and more effectively than iterative message passing. Furthermore, from the node classification accuracy distribution on different SLDs shown in Figure~\ref{fig:SLD}, it can be seen that for a heterophilious graph PCGCN prominently boosts the classification performance on the nodes whose nearest labeled nodes are in 1 or 2-hop neighborhood, i.e., $SLD = 1, 2$, compared to the vanilla GCN. This verifies the rectification effect of pinning control in learning the features via message aggregation. Specifically, the basic message passing aggregates features from the incompatible neighbors, while pining control injects the class-relevant features to the aggregation, preventing the learned representations to stray away from their ground truth classes. 

\subsubsection{Influence of the Labels}
As the prototypes used for feature supervision in PCGCN are derived from the labeled data, it is essential to study the influence of labeled nodes on PCGCN's performance. We evaluate this from two aspects: 1) the labeled data are limited; and 2) some classes are not labeled. For the first case, we vary the size of the training set from $70$ to $2$ on both homophilous and heterophilious datasets, respectively. We average the results over 10 runs on the datasets using random train/validation/test splits for each training set size.  The results are shown in Figure~\ref{fig:label}. It can be observed that under different training size settings, PCGCN (red line) consistently surpasses the baseline models on all datasets with varying number of labeled nodes, suggesting that pinning control enables GCN to exploit both the labeled and structural information with state feedback supervision. In particular, the large margin between PCGCN and the vanilla message passing GNNs on heterophilious graphs (i.e., Chameleon, Squirrel and Actor) indicates again that pinning control is effective to mitigate the heterophily issue.

Since the class prototypes in pinning controllers are defined as the centers of labeled nodes in the same classes, it raises a question: when there is no labeled data for some classes (i.e., Texas and Connell, where some classes of labels are missing from the training set in some splits), how can one derive the corresponding controllers for the nodes of this class? To resolve the robustness of PCGCN in the label missing situation, actually only minor changes are needed: the prototype corresponding to the class with unlabeled nodes is randomly initialized, and further learned in the training phase, as described in line 8 in Algorithm~\ref{alg:algor}. 

We conduct experiments on two heterophilious graphs, i.e., Chameleon and Squirrel, by masking the labels of a certain class. The resultant datasets are denoted as “dataset-i”, where $i$ is the index of that class. We compare PCGCN with the vanilla GCN and GAT, and non-message passing model MLP. From Table~\ref{tb:missclass-1} and Table~\ref{tb:missclass-2} one can observe that the performances of all models are degraded on the datasets with label missing for certain classes, compared to the situations with the original datasets (i.e., the second column in these two tables). However, compared to three baselines, PCGCN preserves good performance on masked datasets, implying that by learning the prototype of missing classes PCGCN is robust against label missing. 

\begin{table*}[htbp]
	\centering
	\renewcommand\arraystretch{1.3}
	\setlength\tabcolsep{5.0pt}
	\caption{Ablation study.}
	\label{Ablation}
	\begin{tabular}{lccccccccc}
		\hline
		\textbf{Datasets} & \textbf{Cora} & \textbf{CiteSeer} & \textbf{Pubmed} & \textbf{Chameleon} & \textbf{Squirrel} & \textbf{Actor}  & \textbf{Connell} & \textbf{Wisconsin} & \textbf{Texas} \\
		\hline
		w/o Hom-P & 87.40 & 76.83 & 89.52 & 73.85 & 65.26 & 35.91& 85.13 & 82.15 & 83.78
		\\
		w/o Het-P & 87.70 & 76.97 &  89.54 & 72.69 & 64.52 & 35.94 & 83.24 & 80.98 & 83.24\\
		w/o MP & 76.53 & 73.20 & 88.87 & 49.86 & 45.41 & 36.48 & 83.78 &  81.37 & 83.24\\
		w/o CL & 87.08 & 76.94 & 89.40 & 64.01 & 54.60 & 35.51 & 85.13 & 84.90 & 82.16\\
		\hline
		PCGCN & \textbf{87.65} & \textbf{77.40} & \textbf{90.34} & \textbf{74.29} & \textbf{65.47} & \textbf{36.43}& \textbf{85.94} & \textbf{87.64} & \textbf{85.95}\\
		\hline
	\end{tabular} 
\end{table*}
\begin{table*}[htbp]
	\centering
	\renewcommand\arraystretch{1.3}
	\setlength\tabcolsep{5.0pt}
	\caption{ Performance comparison between full control and partial control. 10\% of the nodes are selected to be uncontrolled in three different ways: “-Random” denotes random selection, “-MinD.” denotes the selection of the last 10 percent of nodes in terms of node degree, while “-MaxD.” represents the selection of the top 10 percent of nodes also in terms of node degree. }
	\label{controllers}
	\begin{tabular}{lccccccccc}
		\hline
		\textbf{Datasets} & \textbf{Chameleon} & \textbf{Squirrel} & \textbf{Actor} & \textbf{Texas} & \textbf{Wisconsin} & \textbf{Cornell}& \textbf{Cora} & \textbf{Citeseer} & \textbf{Pubmed}\\
		\hline
		
            PCGCN-Random(10\%) & 71.14&	61.97&	33.80 &74.32&	78.82&	79.72&
		87.28&	76.63	&89.22\\
            PCGCN-MinD.(10\%) & 72.25&	62.95&	33.19 &81.62&	75.29&	79.18&
		87.68&	77.24	&90.18\\
            PCGCN-MaxD.(10\%) & 73.57&	64.83&	35.03 &82.70&	85.09&	82.43 &87.40&77.05&89.74\\
		\hline
		PCGCN & \textbf{74.29} & \textbf{65.47} & \textbf{36.43} & \textbf{85.94} & \textbf{87.64} & \textbf{85.95}& \textbf{87.65} & \textbf{77.40} & \textbf{90.34} \\
		\hline
	\end{tabular} \label{tab:cNum}
\end{table*}

\begin{figure}[htpb]
	\centering
	\includegraphics[scale=0.3]{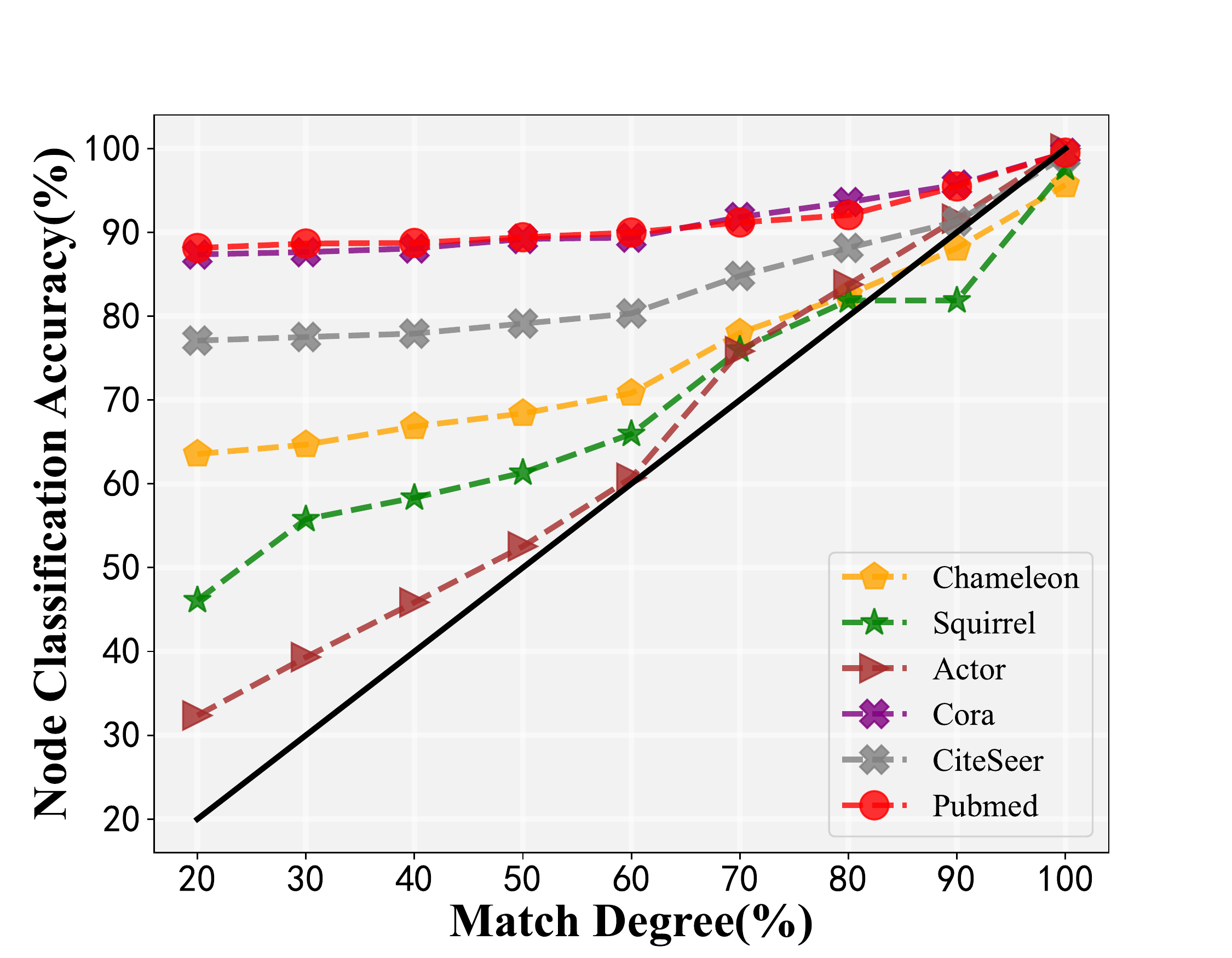}
	\caption{Correlation between matching degree and prediction. The bold black line is the reference of perfect fitting.}
	\label{fig:corr}
\end{figure}

\subsubsection{Correlation between Matching and Prediction}

We explore the relationship between model performance and the degree of matching  between the labels of the “desired states” (i.e., prototypes) and the labels of the unlabeled nodes. To make the match degree be a controllable variable in the experiments, we set a certain percentage of unlabeled nodes with their ground-truth labels in pinning control matching matrix $B$ and then train PCGCN. The results in Figure~\ref{fig:corr} show that the performance of PCGCN is strongly correlated to the matching degree on heterophilious graphs while less correlated on homophilous graphs, indicating the importance of learning a good matching relation between prototypes and nodes to resolve the heterophily issue.

\subsubsection{Full Control vs. Partial Control}
We have compared the performances of PCGCN under two control schemes, namely, full control and partial control. In the experiments, we implemented partial control by retaining 10\% of the nodes uncontrolled. We also consider three different ways to select the uncontrolled nodes to retain: random retaining, top 10\% nodes retaining and last 10\% nodes retaining, in terms of node degrees. Results shown in Table~\ref{tab:cNum} suggest that full control is necessary to achieve the best performance, compared to the other partial control schemes. The reason is that pinning controllers serve as the supervisors to enhance class-relevant feature learning. It is also noteworthy that top 10\% retaining is superior to the other two schemes on all datasets with strong heterophily, but slightly inferior to the other two on the datasets with strong homophily (i.e., Cora, Citeseer and Pubmed). This reflects that high-degree nodes are less affected by heterophily in heterophilous graphs than in homophilious graphs.

\subsection{Ablation Study}
To evaluate the effects of different pinning relation propagation methods (i.e., homophily propagation (Ho-P) and heterophily propagation (He-P) corresponding to the first and second terms in Eq.(\ref{con:sim})), message passing (MP) and consistency loss (CL) for pinning control, we conduct ablation experiments on homophily and heterophily datasets, respectively.  The first and second rows of Table~\ref{Ablation} report the results of PCGCN without homophily pinning similarity propagation and those without heterophily pinning similarity propagation. It can be observed that for graphs with strong homophily (i.e., Cora, CiteSeer and Pubmed), two propagation schemes have little influence on model performance. In contrast, these two schemes play a role on heterophilious graphs. %we found that pinning relation propagation manner has less effect on the model than MP and CL, and that removing H-similarity on homophily graph also had a greater effect and removing L-similarity on heterophily graph also had a greater effect. 
The third row of the table lists the results by replacing message passing in Eq.(\ref{eq:hybrid}) with the feature representations of the previous layer, while in the fourth row of the Table, the consistency loss that penalizes the pinning control matching degree is removed from the training model. Experimental results suggest that both structural information (captured by message passing) and control matching (regularized by consistency loss) have  significant impacts on improving the performance.

\section{Conclusion}
In this paper, to address the challenges from limited training samples and heterophilious graphs, we propose a pinning control scheme for boosting message passing GNNs from the control theoretic viewpoint. By assuming the prototypes of labeled data to be the desired representations for different types of nodes, we integrate the state feedback control into the vanilla message passing to achieve prototype-supervised graph representation learning. The experiments on homophilous and heterophilious benchmark graphs show that the proposed PCGCN brings great gains to the vanilla GCN and outperforms the comparable leading GNNs. PCGCN enables us to explore the supervision ability of labeled data. Our research will pave the way for devising more effective supervision enhancement techniques in the future.

\section*{Acknowledgments}
{This work is supported by the National Natural Science Foundation (NSFC 62276099) and SWPU Innovation Base funding (No.642).}

\bibliographystyle{IEEEtran}
\bibliography{reference}

\end{document}